\documentclass[sigconf,authorversion]{aamas}

\renewcommand\footnotetextcopyrightpermission[1]{}
\usepackage{balance} % for balancing columns on the final page
\usepackage{multirow} % for multi-row cells in tables
\usepackage{rotating} % for rotating text
\usepackage{booktabs} % for enhanced table formatting
\usepackage{subcaption} % for subtables and subfigures
\usepackage{colortbl} % for colored table cells
\usepackage{xcolor} % for colors
\usepackage{algorithm} % for algorithm environment
\usepackage{algorithmic} % for algorithmic pseudocode

\definecolor{lightblue}{RGB}{200,220,255} % light blue for our methods (enhanced)
\definecolor{lightgreen}{RGB}{220,255,220} % light green for key equations (enhanced)
\definecolor{modulecolor}{RGB}{0,100,255} % module color (vibrant blue for better visibility)
\definecolor{highlight}{RGB}{255,100,0} % bright orange for important numbers (enhanced visibility)

\newcommand{\module}[1]{\textcolor{modulecolor}{\textbf{#1}}}
\newcommand{\AuthorGuowei}{Guowei Zou}
\newcommand{\AuthorHaitao}{Haitao Wang}
\newcommand{\AuthorHejun}{Hejun Wu}
\newcommand{\AuthorYukun}{Yukun Qian}
\newcommand{\AuthorYuhang}{Yuhang Wang}
\newcommand{\AuthorWeibing}{Weibing Li}
\newcommand{\AuthorPdfList}{\AuthorGuowei, \AuthorHaitao, \AuthorHejun, \AuthorYukun, \AuthorYuhang, \AuthorWeibing}
\newcommand{\AuthorSupGuowei}{\AuthorGuowei$^{1,2,*}$}
\newcommand{\AuthorSupHaitao}{\AuthorHaitao$^{1,2,*}$}
\newcommand{\AuthorSupHejun}{\AuthorHejun$^{1,2}$}
\newcommand{\AuthorSupYukun}{\AuthorYukun$^{1,2}$}
\newcommand{\AuthorSupYuhang}{\AuthorYuhang$^{1,2}$}
\newcommand{\AuthorSupWeibing}{\AuthorWeibing$^{1,2,\dagger}$}
\newcommand{\AffilOneText}{$^{1}$School of Computer Science and Engineering, Sun Yat-sen University, Guangzhou, China}
\newcommand{\AffilTwoText}{$^{2}$Guangdong Key Laboratory of Big Data Analysis and Processing, Guangzhou, China}

\newcommand{\MailListMailTwo}{\{zougw, wanght39, qianyk5, wangyh253\}@mail2.sysu.edu.cn}
\newcommand{\MailListMail}{\{wuhejun, liwb53\}@mail.sysu.edu.cn}

\newcommand{\EmailGuowei}{zougw@mail2.sysu.edu.cn}
\newcommand{\EmailHaitao}{wanght39@mail2.sysu.edu.cn}
\newcommand{\EmailHejun}{wuhejun@mail.sysu.edu.cn}
\newcommand{\EmailYukun}{qianyk5@mail2.sysu.edu.cn}
\newcommand{\EmailYuhang}{wangyh253@mail2.sysu.edu.cn}
\newcommand{\EmailWeibing}{liwb53@mail.sysu.edu.cn}
\newcommand{\affilSYSU}{%
  \affiliation{%
    \institution{School of Computer Science and Engineering, Sun Yat-sen University}
    \city{Guangzhou}
    \state{Guangdong}
    \country{China}}}
\newcommand{\affilLab}{%
  \affiliation{%
    \institution{Guangdong Key Laboratory of Big Data Analysis and Processing}
    \city{Guangzhou}
    \state{Guangdong}
    \country{China}}}
\newcommand{\customauthorblock}{%
  \begin{center}
    \large \textbf{\AuthorSupGuowei\quad \AuthorSupHaitao\quad \AuthorSupHejun\quad \AuthorSupYukun\quad \AuthorSupYuhang\quad \AuthorSupWeibing}\\
    \vspace{0.3em}
    \normalsize \AffilOneText\\
    \normalsize \AffilTwoText\\
    \vspace{0.2em}
    \normalsize Emails: \MailListMailTwo; \MailListMail
    \vspace{0.15em}
  \end{center}}

\hypersetup{pdfauthor={\AuthorPdfList}}

\makeatletter
\def\@mkauthors@iii{%
  \global\setbox\mktitle@bx=\vbox{\noindent
    \unvbox\mktitle@bx\par\medskip
    \customauthorblock
    \par\bigskip}}
\def\@authornotes{}
\def\authornote#1{%
  \if@ACM@anonymous\else
    \g@addto@macro\@authornotes{%
      \footnotetext[\z@]{#1}}%
  \fi}
\makeatother

\setcopyright{ifaamas}
\acmConference[AAMAS '26]{Proc.\@ of the 25th International Conference
on Autonomous Agents and Multiagent Systems (AAMAS 2026)}{May 25 -- 29, 2026}
{Paphos, Cyprus}{C.~Amato, L.~Dennis, V.~Mascardi, J.~Thangarajah (eds.)}
\copyrightyear{2026}
\acmYear{2026}
\acmDOI{}
\acmPrice{}
\acmISBN{}

\acmSubmissionID{<<submission id>>}

\title[DM1: MeanFlow with Dispersive Regularization]{DM1: MeanFlow with Dispersive Regularization for 1-Step Robotic Manipulation}

\author{\AuthorGuowei}
\affilSYSU
\affilLab
\email{\EmailGuowei}

\author{\AuthorHaitao}
\affilSYSU
\affilLab
\email{\EmailHaitao}

\author{\AuthorHejun}
\affilSYSU
\affilLab
\email{\EmailHejun}

\author{\AuthorYukun}
\affilSYSU
\affilLab
\email{\EmailYukun}

\author{\AuthorYuhang}
\affilSYSU
\affilLab
\email{\EmailYuhang}

\author{\AuthorWeibing}
\affilSYSU
\affilLab
\email{\EmailWeibing}
\authornote{*Equal contribution. \dag Corresponding author.}

\begin{abstract}
The ability to learn multi-modal action distributions is indispensable for robotic manipulation policies to perform precise and robust control. Flow-based generative models have recently emerged as a promising solution to learning distributions of actions, offering one-step action generation and thus achieving much higher sampling efficiency compared to diffusion-based methods. However, existing flow-based policies suffer from representation collapse, the inability to distinguish similar visual representations, leading to failures in precise manipulation tasks.  
We propose \textbf{DM1} (MeanFlow with Dispersive Regularization for One-Step Robotic Manipulation), a novel flow matching framework that integrates dispersive regularization into MeanFlow to prevent collapse while maintaining one-step efficiency. DM1 employs multiple dispersive regularization variants across different intermediate embedding layers, encouraging diverse representations across training batches without introducing additional network modules or specialized training procedures.    Experiments on RoboMimic benchmarks show that DM1 achieves 20--40$\times$ faster inference (0.07s vs.\ 2--3.5s) and improves success rates by 10--20 percentage points, with the Lift task reaching 99\% success over 85\% of the baseline. Physical deployment on a Franka-Emika-Panda robot further validates that DM1 transfers effectively from simulation to the real world. To the best of our knowledge, this is the first work to leverage representation regularization to enable flow-based policies to achieve strong performance in robotic manipulation, establishing a simple yet powerful approach for efficient and robust manipulation. \textbf{Website with code: \href{https://guowei-zou.github.io/dm1/}{\textcolor{blue}{https://guowei-zou.github.io/dm1/}}}
\end{abstract}

\newcommand{\BibTeX}{\rm B\kern-.05em{\sc i\kern-.025em b}\kern-.08em\TeX}

\begin{document}

%%% The following commands remove the headers in your paper. For final 
%%% papers, these will be inserted during the pagination process.

\pagestyle{fancy}
\fancyhead{}

%%% The next command prints the information defined in the preamble.

\maketitle

\section{Introduction}

\begin{figure}[h]
\centering
\includegraphics[width=0.48\textwidth]{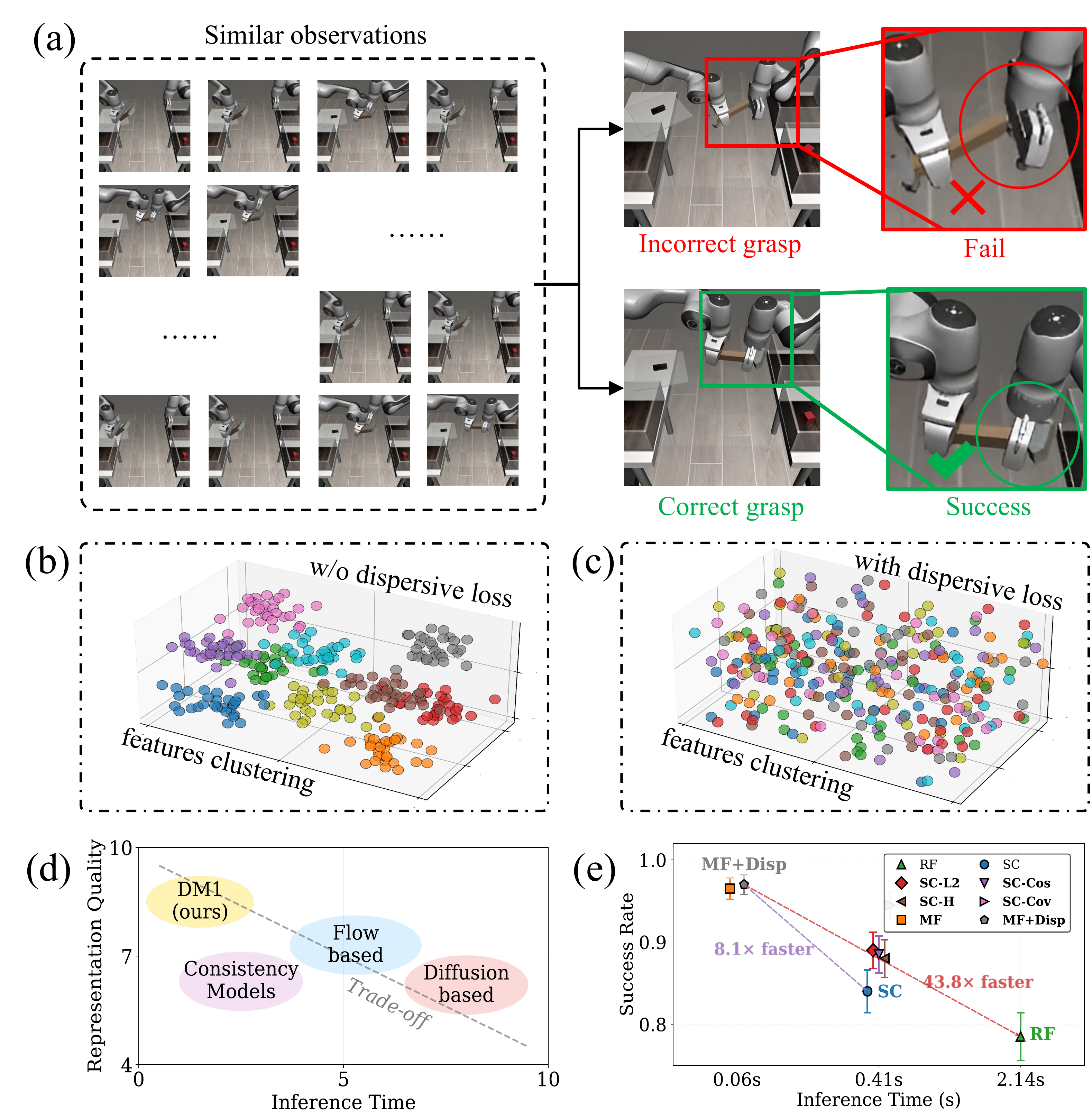}
\caption{Visualization of the effect of dispersive regularization in DM1.
(a) Example rollouts showing how similar observations can lead to incorrect vs. correct grasps; (b–c) Feature distributions without and with dispersive loss, where dispersion prevents representation collapse; (d) Method landscape illustrating the speed–quality trade-off; and (e) Quantitative comparison of success rate versus inference time, showing DM1’s superior efficiency and accuracy.}
\label{fig:motivation}
\end{figure}

Robotic manipulation requires control policies to be capable of capturing and utilizing multi-modal action distributions in addition to satisfying real-time control constraints. Generative controllers built on diffusion and flow-based models \cite{dppo2023,dp3_2023,reinflow2024,diffusion_policy_survey2025} leverage the modeling power of denoising diffusion probabilistic models \cite{ddpm2020,ddim2020,score2019,ncsn2020} and continuous normalizing flows \cite{cnf2018,lipman2023flow,rectifiedflow2022}, but practical deployment remains challenging due to the difficulty of achieving both expressiveness and low latency.

As illustrated in Figure~\ref{fig:motivation}(d), existing methods face a fundamental trade-off between speed and representation quality. Diffusion-style policies \cite{dppo2023,dp3_2023} and related generative modeling approaches in vision \cite{ldm2022,dit2023,edm2022} can model expressive multi-modal behaviors, but require 50--100 neural function evaluations (NFEs, essentially 50--100 forward passes) per control step, precluding real-time control. Flow-based models \cite{reinflow2024,flowpolicy2024,affordance_flow2024}, which learn continuous transformations between noise and data distributions, reduce the number of sampling steps but remain dependent on iterative Ordinary Differential Equation (ODE) integration. To achieve one-step generation, beyond these two families, consistency-based approaches \cite{consistency2023} propose an alternative framework but require specialized training and additional network modules. In parallel, within the flow matching family, mean velocity methods (e.g., MeanFlow \cite{meanflow2024}) and characteristic learning \cite{characteristic_onestep2025} achieve one-NFE efficiency by directly predicting the mean velocity of the action trajectory, avoiding iterative ODE solving. However, this computational efficiency comes at a critical cost: these mean-velocity methods employ tighter mathematical constraints that cause distinct observations to map to overly similar embeddings, a phenomenon known as \textit{representation collapse}, as shown in Fig.~\ref{fig:motivation}(b). This collapse limits the capacity to capture diverse multi-modal behaviors and is particularly detrimental to fine-grained manipulation tasks requiring discrimination of subtle variations, such as grasping objects at different orientations, as depicted in Fig.~\ref{fig:motivation}(a).

Addressing this collapse without sacrificing one-step efficiency presents a fundamental challenge. The root cause lies in the training objective itself: the mean-velocity formulation does not explicitly encourage diversity among learned representations. This observation suggests enforcing feature separation at intermediate representation layers during training, as presented in Fig.~\ref{fig:motivation}(c). By regularizing embeddings within each batch to maintain sufficient distance in latent space, collapse can be prevented at its source without additional architectural components.

To this end, this paper proposes DM1 (MeanFlow with Dispersive Regularization for One-Step Robotic Manipulation), which integrates dispersive regularization into MeanFlow to maintain one-NFE efficiency while preventing representation collapse. 

As summarized in Figure~\ref{fig:motivation}(e), the main contributions are: 

\textbf{(1) Breaking the speed-accuracy trade-off.} DM1 achieves 20--40$\times$ faster inference than baseline methods while improving success rates by 10--20 percentage points on complex tasks. With only 1 step, DM1 attains competitive performance across manipulation tasks of varying difficulty, eliminating the traditional compromise between latency and expressiveness.

\textbf{(2) Preventing representation collapse in one-step generation.} DM1 represents the first systematic integration of dispersive regularization with MeanFlow for vision-based robotic manipulation. The approach applies dispersive losses to multiple intermediate embedding layers without architectural modifications, addressing collapse at its source while preserving one-step generation efficiency.

\textbf{(3) Systematic evaluation of dispersive regularization strategies.} This work provides detailed analysis of four dispersive regularization variants (InfoNCE-L2, InfoNCE-Cosine, Hinge, Covariance-based) across varying task complexity and regularization weight configurations, demonstrating their effectiveness in preventing collapse when applied to intermediate representations.

\textbf{(4) Real-world deployment validation.} Comprehensive experiments on RoboMimic benchmarks across multiple denoising step configurations demonstrate consistent performance gains. Physical deployment on a Franka-Emika-Panda robot validates practical applicability, achieving latencies that enable real-time control at frequencies exceeding 50Hz.

The remainder of this paper is organized as follows: Section 2 reviews related work in flow-based robotic control and representation learning; Section 3 presents the DM1 methodology and theoretical foundations; Section 4 reports experimental results on benchmark tasks and physical robot validation; and Section 5 concludes with discussions and future directions.

\section{Related Work}

Recent surveys \cite{diffusion_survey2025,diffusion_manipulation_survey2025,vla_survey2025} have reviewed robot manipulation and vision-language-action model advances. This work focuses on efficient one-step generation while addressing representation collapse.

\subsection{Flow-Based Models for Robotic Control}

Flow-based generative models have gained attention in robotic control for modeling complex, multimodal action distributions. Building upon continuous normalizing flows \cite{cnf2018}, flow matching \cite{lipman2023flow} simplified this paradigm by directly regressing velocity fields, while Rectified Flow \cite{rectifiedflow2022} improved efficiency by learning straight trajectories in probability space. Diffusion Policy \cite{dppo2023} pioneered diffusion models for visuomotor control, with extensions like DP3 \cite{dp3_2023} handling 3D point cloud observations. Recent one-step diffusion approaches include 1-DP \cite{onedp2025}. Flow-based methods like FlowPolicy \cite{flowpolicy2024}, ReinFlow \cite{reinflow2024}, recent shortcut flows \cite{shortcutflow2025}, and concurrent work on PointFlowMatch \cite{pointflowmatch2024} and FlowRAM \cite{flowram2025} have reduced inference steps but still require 20--100 NFEs, creating computational overhead that limits real-time deployment. Large-scale foundation models including $\pi_{0.5}$ \cite{pi05_2024}, $\pi_0$ \cite{pi0_2024}, and GraspVLA \cite{graspvla2024} employ flow matching in expert architectures, yet the fundamental challenge of achieving true one-step generation while maintaining expressiveness remains unsolved. MeanFlow \cite{meanflow2024} enables one-step generation via average velocity learning. MP1 \cite{mp1_meanflow} applied it to robotic manipulation, achieving 1-NFE inference. However, \textit{representation collapse} limits its ability to capture behavioral diversity. This work addresses this via dispersive regularization while maintaining efficiency.

\subsection{Representation Learning and Dispersive Regularization}

Dispersive regularization techniques encourage internal representations to disperse in the hidden space, addressing representation collapse in deep learning models. These approaches draw inspiration from contrastive learning frameworks such as SimCLR \cite{simclr2020} and self-supervised methods like Barlow Twins \cite{barlow2021} and VICReg \cite{vicreg2022}, which operate without requiring positive pairs or data augmentation. Related work in representation learning includes methods like R3M \cite{r3m2022}, MVP \cite{mvp2022}, and visual pre-training approaches using CLIP \cite{clip2021} and Perceiver \cite{perceiver2021}. Recent work has explored dispersive regularization for image generation \cite{diffuse_and_disperse} and diffusion policies \cite{d2ppo}, introducing dispersive loss to address representation collapse by treating hidden representations within batches as negative pairs to enforce discriminative embeddings. Various formulations exist, including InfoNCE-based approaches, hinge losses, and covariance-based methods. This work builds upon this foundation by incorporating dispersive regularization into MeanFlow, achieving both computational efficiency and robust representations.

\begin{figure*}[t]
\centering
\includegraphics[width=0.95\textwidth]{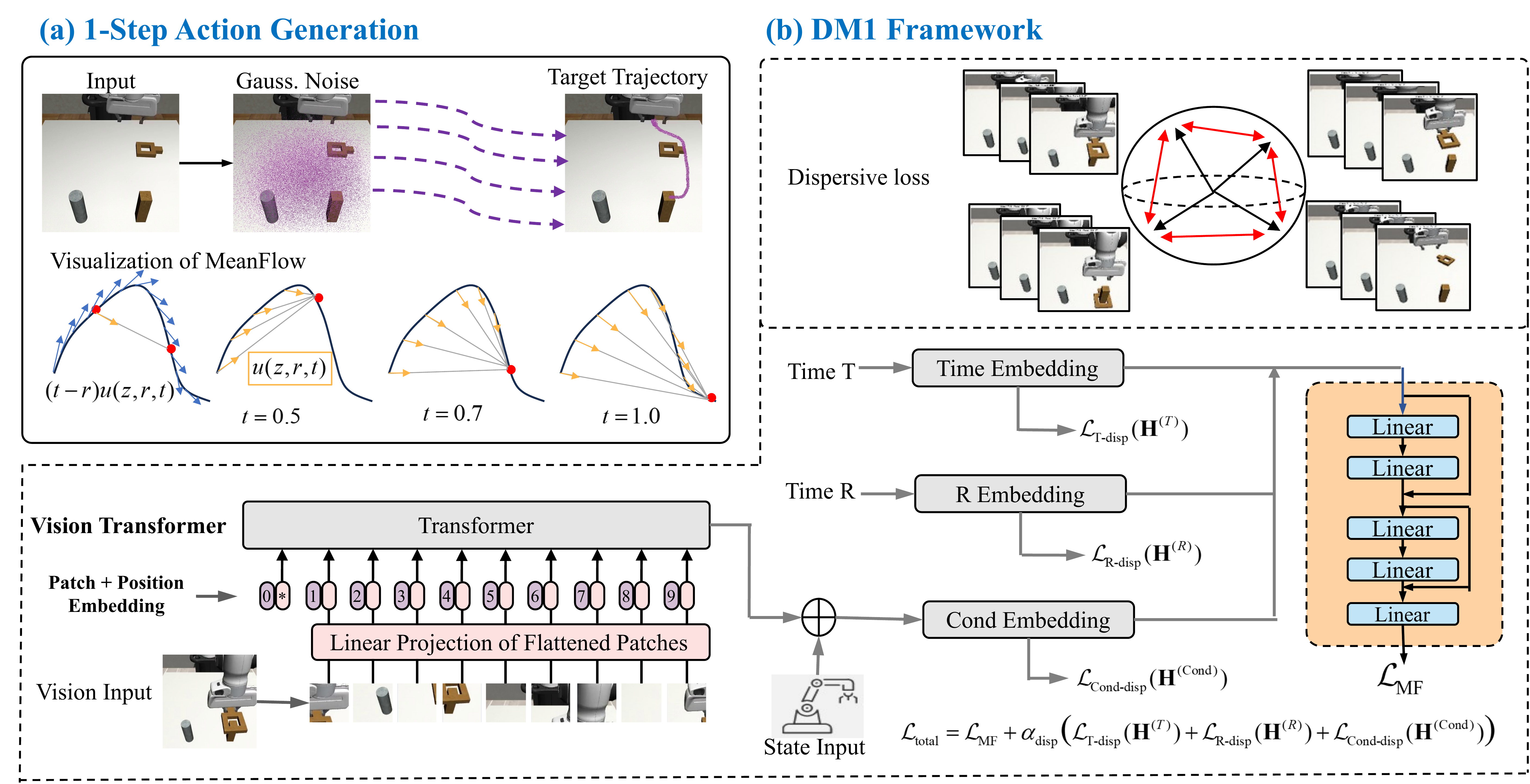}
\caption{DM1 Framework Architecture. \textbf{Top Left:} 1-Step Action Generation showing MeanFlow's core principle of direct trajectory generation through average velocity fields, contrasting with traditional multi-step denoising approaches. \textbf{Bottom Left:} Vision Transformer Encoder processing input images into patch tokens with positional encoding for global visual feature extraction. \textbf{Top Right:} Dispersive Loss components (R Disp., T Disp., Cond Disp.) encouraging embedding separation across different modalities to prevent representation collapse. \textbf{Bottom Right:} Complete DM1 computational flow integrating vision input, state input, and temporal conditions through embedding modules, with dispersive losses applied to intermediate representations and MeanFlow loss for velocity field prediction.}
\label{fig:dm1_framework}
\end{figure*}

\section{Methodology}

\subsection{Framework Overview}

Figure \ref{fig:dm1_framework} presents the DM1 framework, which integrates MeanFlow one-step generation with dispersive regularization across multimodal inputs. The framework comprises four key components:

\textbf{(1)} \module{One-Step Action Generation:} Gaussian noise is directly transformed into target action trajectories through learned average velocity fields, eliminating iterative denoising. This achieves high-quality action synthesis in a single NFE.

\textbf{(2)} \module{Vision Transformer Encoder:} Visual observations are processed through a patch-based Vision Transformer (ViT) architecture to extract global visual features that serve as conditional inputs for the flow matching model. Alternative architectures like ResNet \cite{resnet2016} could be used but ViT provides better global context modeling.

\textbf{(3)} \module{Dispersive Loss Modules:} Our key contribution for addressing representation collapse. Dispersive regularization is applied across multiple embedding layers (R, T, and Cond embeddings) to enforce representational diversity and prevent collapse.

\textbf{(4)} \module{Complete DM1 Architecture:} The integrated pipeline combines vision encoder outputs, proprioceptive state inputs, and temporal embeddings. The system is trained using MeanFlow loss for velocity field learning and dispersive losses for regularization, ensuring both computational efficiency and robust representation learning.

%%%%%%%%%%%%%%%%%%%%%%%%%%%%%%%%%%%%%%%%%%%%%%%%%%%%%%%%%%%%%%%%%%%%%%%%

\subsection{Flow Matching Preliminaries}

Flow matching provides a framework for learning transformations between distributions through continuous normalizing flows with dynamics:
\begin{equation}
    \frac{dz_t}{dt} = v_t(z_t),
\end{equation}
where $z_t$ is the trajectory state at time $t \in [0,1]$ and $v_t(z_t)$ is the velocity field.

For tractable training, we use linear interpolation:
\begin{equation}
    z_t = (1-t)x_0 + tx_1,
\end{equation}
where $x_0 \sim p_0$ is the initial noise distribution, $x_1 \sim p_1$ is the target data distribution, and $t \in [0,1]$ is the interpolation parameter.

This yields the conditional velocity field:
\begin{equation}
    v_t(z_t | x_0, x_1) = x_1 - x_0,
\end{equation}
which serves as the training target.

\subsection{MeanFlow Training Objective}

We follow the standard Rectified Flow formulation with linear interpolation:
\begin{equation}
    z_t = (1-t)\epsilon + ta,
\end{equation}
where $\epsilon \sim \mathcal{N}(0,I)$ is a Gaussian noise, $a \sim p(a)$ is the target action trajectory sampled from the data distribution, $z_t \in \mathbb{R}^{T_a \times D_a}$ represents the trajectory state at time $t \in [0,1]$, with $T_a$ denoting the action horizon (number of time steps) and $D_a$ the action dimensionality.

The conditional velocity field is:
\begin{equation}
    v_t(z_t \mid \epsilon,a) = a - \epsilon,
\end{equation}
which defines the direction from noise to data (we adopt this sign convention throughout this paper).

The \emph{average velocity field} is defined for $r < t$ as
\begin{equation}
    u(z_t,r,t) \triangleq \frac{1}{t-r}\int_r^t v(z_\tau,\tau)\, d\tau,
\end{equation}
where $u(z_t,r,t) \in \mathbb{R}^{T_a \times D_a}$ represents the average velocity over the time interval $[r,t]$ with $r,t \in [0,1]$ and $r < t$. This formulation satisfies the displacement identity $(t-r)u(z_t,r,t) = z_t - z_r$.

Differentiating the displacement yields the \textbf{MeanFlow identity}:
\begin{equation}
    u(z_t,r,t) = v(z_t,t) - (t-r)\frac{d}{dt}u(z_t,r,t).
\end{equation}
The total derivative expands to
\begin{equation}
    \frac{d}{dt}u(z_t,r,t) = v(z_t,t)\cdot \nabla_{z_t}u(z_t,r,t) + \partial_t u(z_t,r,t),
\end{equation}
where the $\cdot$ denotes a Jacobian–vector product (JVP) with vector $v(z_t,t)=a-\epsilon$.

We thus obtain the MeanFlow training objective:
\begin{equation}
\mathcal{L}_{\text{MF}} = \mathbb{E}_{a,\epsilon,t,r} \left[
\big\| u_\theta(z_t,r,t,o_t) - \mathrm{sg}(u_{\text{tgt}}) \big\|_2^2
\right],
\end{equation}
where $u_\theta(z_t,r,t,o_t)$ is the neural network parameterized by $\theta$ that predicts the average velocity given trajectory state $z_t$ at time $t$, time parameters $r,t \in [0,1]$ with $r < t$, and observation $o_t \in \mathbb{R}^{D_o}$ (including visual and proprioceptive features with dimensionality $D_o$). The operator $\mathrm{sg}(\cdot)$ denotes stop-gradient to prevent gradients from flowing through the target.

The target $u_{\text{tgt}}$ is computed as:
\begin{equation}
u_{\text{tgt}} = (a-\epsilon) - (t-r)\left((a-\epsilon)\cdot \nabla_{z_t}u_\theta + \partial_t u_\theta \right).
\end{equation}

\subsection{Dispersive Regularization for Preventing Collapse}

MeanFlow suffers from \textit{representation collapse} where distinct observations map to similar embeddings. Dispersive regularization addresses this by encouraging intermediate features $\mathbf{H} = \{\mathbf{h}_1, \ldots, \mathbf{h}_B\} \subset \mathbb{R}^d$ to spread in feature space, where $\mathbf{h}_i \in \mathbb{R}^d$ is the $i$-th feature vector in a batch of size $B$ (number of samples per training batch) and $d$ is the feature dimension of the intermediate representation layer. We explore four formulations:

\textbf{InfoNCE-L2:} Maximizes pairwise Euclidean distances using contrastive learning:

\begin{equation}
\mathcal{L}_{\text{InfoNCE-L2}} = -\mathbb{E}_{\mathbf{h}_i, \mathbf{h}_j \sim \mathcal{B}} \left[ \log \frac{\exp(-\|\mathbf{h}_i - \mathbf{h}_j\|_2^2 / (2\tau^2))}{\sum_{k \neq i} \exp(-\|\mathbf{h}_i - \mathbf{h}_k\|_2^2 / (2\tau^2))} \right]
\end{equation}

\noindent where $\mathbf{h}_i, \mathbf{h}_j \in \mathbb{R}^d$ are feature vectors sampled from the batch distribution $\mathcal{B}$, $\tau > 0$ is the temperature parameter controlling concentration, and $\|\cdot\|_2$ denotes the Euclidean norm.

\textbf{InfoNCE-Cosine:} Maximizes angular diversity between feature directions:

\begin{equation}
\mathcal{L}_{\text{InfoNCE-Cos}} = -\mathbb{E}_{\mathbf{h}_i, \mathbf{h}_j \sim \mathcal{B}} \left[ \log \frac{\exp(\mathbf{h}_i^T \mathbf{h}_j / (\|\mathbf{h}_i\|_2 \|\mathbf{h}_j\|_2 \tau))}{\sum_{k \neq i} \exp(\mathbf{h}_i^T \mathbf{h}_k / (\|\mathbf{h}_i\|_2 \|\mathbf{h}_k\|_2 \tau))} \right]
\end{equation}

\noindent where $\mathbf{h}_i^T \mathbf{h}_j$ denotes the dot product between feature vectors, and the denominator normalizes by feature magnitudes to compute cosine similarity.

\textbf{Hinge Loss:} Enforces a minimum separation margin between representations:

\begin{equation}
\mathcal{L}_{\text{Hinge}} = \mathbb{E}_{\mathbf{h}_i, \mathbf{h}_j \sim \mathcal{B}} \left[ \max(0, \delta - \|\mathbf{h}_i - \mathbf{h}_j\|_2) \right]
\end{equation}

\noindent where $\delta > 0$ is the minimum desired distance between representations, and the loss is activated when the distance falls below $\delta$.

\textbf{Covariance-Based:} Encourages decorrelation across feature dimensions and maintains variance:

\begin{equation}
\mathcal{L}_{\text{Cov}} = \|C - \mathrm{diag}(C)\|_F^2 + \lambda_{\text{cov}} \sum_{i=1}^d \max(0, \sigma_{\min} - C_{ii}), \quad C = \mathrm{Cov}(\mathbf{H})
\end{equation}

\noindent where $C \in \mathbb{R}^{d \times d}$ is the covariance matrix of the feature batch $\mathbf{H}$, $\mathrm{diag}(\cdot)$ extracts the diagonal elements, $\|\cdot\|_F$ denotes the Frobenius norm, $\lambda_{\text{cov}} > 0$ is the weight for variance regularization, $\sigma_{\min} > 0$ is the minimum desired variance for each feature dimension, and $C_{ii}$ is the $i$-th diagonal element representing the variance of the $i$-th feature dimension.

The final training objective combines MeanFlow and dispersive regularization. In its general form:
\begin{equation}
\mathcal{L}_{\text{total}} = \mathcal{L}_{\text{MF}} + \alpha_{\text{disp}} \mathcal{L}_{\text{disp}}(\mathbf{H}^{(l)}),
\end{equation}
where $\mathcal{L}_{\text{MF}}$ is the MeanFlow loss for velocity field learning, $\mathcal{L}_{\text{disp}}$ is one of the four dispersive loss formulations (InfoNCE-L2, InfoNCE-Cosine, Hinge, or Covariance-Based), $\mathbf{H}^{(l)} \in \mathbb{R}^{B \times d}$ denotes the intermediate representation extracted from layer $l$ of the network, and $\alpha_{\text{disp}} > 0$ is the balancing coefficient controlling the strength of dispersive regularization.

In our DM1 framework, we apply dispersive regularization across multiple embedding layers to maximize representational diversity. Specifically, we enforce dispersion on the temporal embedding $\mathbf{H}^{(T)}$, noise embedding $\mathbf{H}^{(R)}$, and conditional embedding $\mathbf{H}^{(\text{Cond})}$:
\begin{equation}
\begin{aligned}
\mathcal{L}_{\text{total}} = \mathcal{L}_{\text{MF}} + \alpha_{\text{disp}} \big[
&\mathcal{L}_{\text{disp}}(\mathbf{H}^{(T)}) \\
&+ \mathcal{L}_{\text{disp}}(\mathbf{H}^{(R)}) + \mathcal{L}_{\text{disp}}(\mathbf{H}^{(\text{Cond})}) \big]
\end{aligned}
\end{equation}
This multi-layer regularization strategy prevents collapse at multiple stages of the representation pipeline, ensuring robust feature diversity throughout the network. The detailed training and inference algorithms are provided in Appendix~\ref{sec:algorithms}.

\begin{figure*}[t]
\centering
\includegraphics[width=0.95\textwidth]{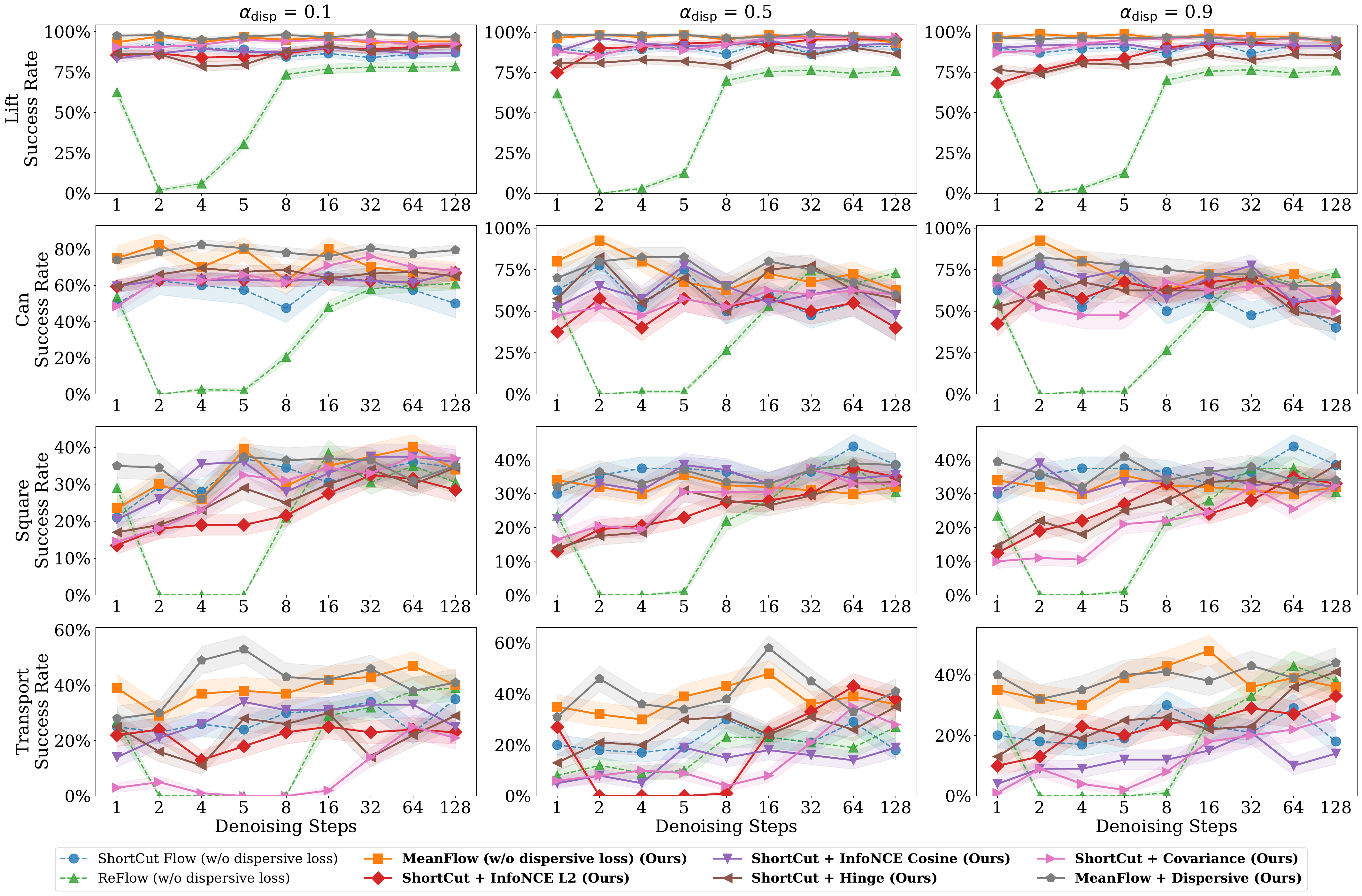}
\caption{Comprehensive evaluation of success rates across varying denoising steps for different weight configurations ($\alpha_{\text{disp}} = 0.1, 0.5, 0.9$) and four robotic manipulation tasks. Each row represents a specific task (Lift, Can, Square, Transport) while columns show different weight factors. The analysis demonstrates the superior performance of our MeanFlow-based approaches (MF, MF+Disp) which achieve competitive success rates with significantly fewer denoising steps (5 steps) compared to baseline methods requiring 32--128 steps.}
\label{fig:evaluation_overview}
\end{figure*}

\section{Experiments}

To comprehensively evaluate the performance and effectiveness of the DM1 framework, our analysis focuses on the following four research questions (RQs):

\noindent\textcolor{red}{\textbf{RQ1: Does DM1 achieve one-step generation efficiency?}}

\noindent\textcolor{red}{\textbf{RQ2: Does dispersive regularization prevent collapse?}}

\noindent\textcolor{red}{\textbf{RQ3: How do different dispersive variants compare?}}

\noindent\textcolor{red}{\textbf{RQ4: Does DM1 transfer to real-world robotic systems?}} 

\subsection{Experimental Setup}

We evaluate DM1 on four robotic manipulation tasks from RoboMimic \cite{robomimic2021}: lift, can, square, and transport, spanning basic grasping to complex multi-stage sequences. We compare against baseline methods: ReFlow (RF, 128 steps) \cite{rectifiedflow2022}, ShortCut Flow (SC, 32 steps), their dispersive variants (SC-L2, SC-Cos, SC-H, SC-Cov), and vanilla MeanFlow (MF) \cite{meanflow2024}. All experiments use MuJoCo \cite{mujoco2012} with a Vision Transformer (ViT) encoder \cite{vit2020} on NVIDIA RTX 4090 Graphics Processing Units (GPUs). Each method is evaluated over 100 episodes per task. See Appendix C for details.

\subsection{Simulation Results}

We present comprehensive results on the RoboMimic benchmark across four critical analysis dimensions, each evaluated through dedicated visualizations and metrics.

\subsubsection{\textbf{Step-wise Performance Analysis (RQ1 \& RQ2)}}

Figure \ref{fig:evaluation_overview} analyzes success rates across varying denoising steps (1, 2, 4, 8, 16, 32, 64, 128) for different weight configurations ($\alpha_{\text{disp}} = 0.1, 0.5, 0.9$) and four tasks, revealing insights into computational efficiency (RQ1) and dispersive regularization effectiveness (RQ2).

Computational Efficiency Advantage: MeanFlow-based methods demonstrate dramatic efficiency gains. While baselines (ReFlow and ShortCut Flow) require 32--128 steps for stable performance, our MeanFlow variants achieve comparable or superior success rates with only 5 steps, representing a 6--25$\times$ reduction in inference steps.

\begin{figure*}[t]
\centering
\includegraphics[width=0.95\textwidth]{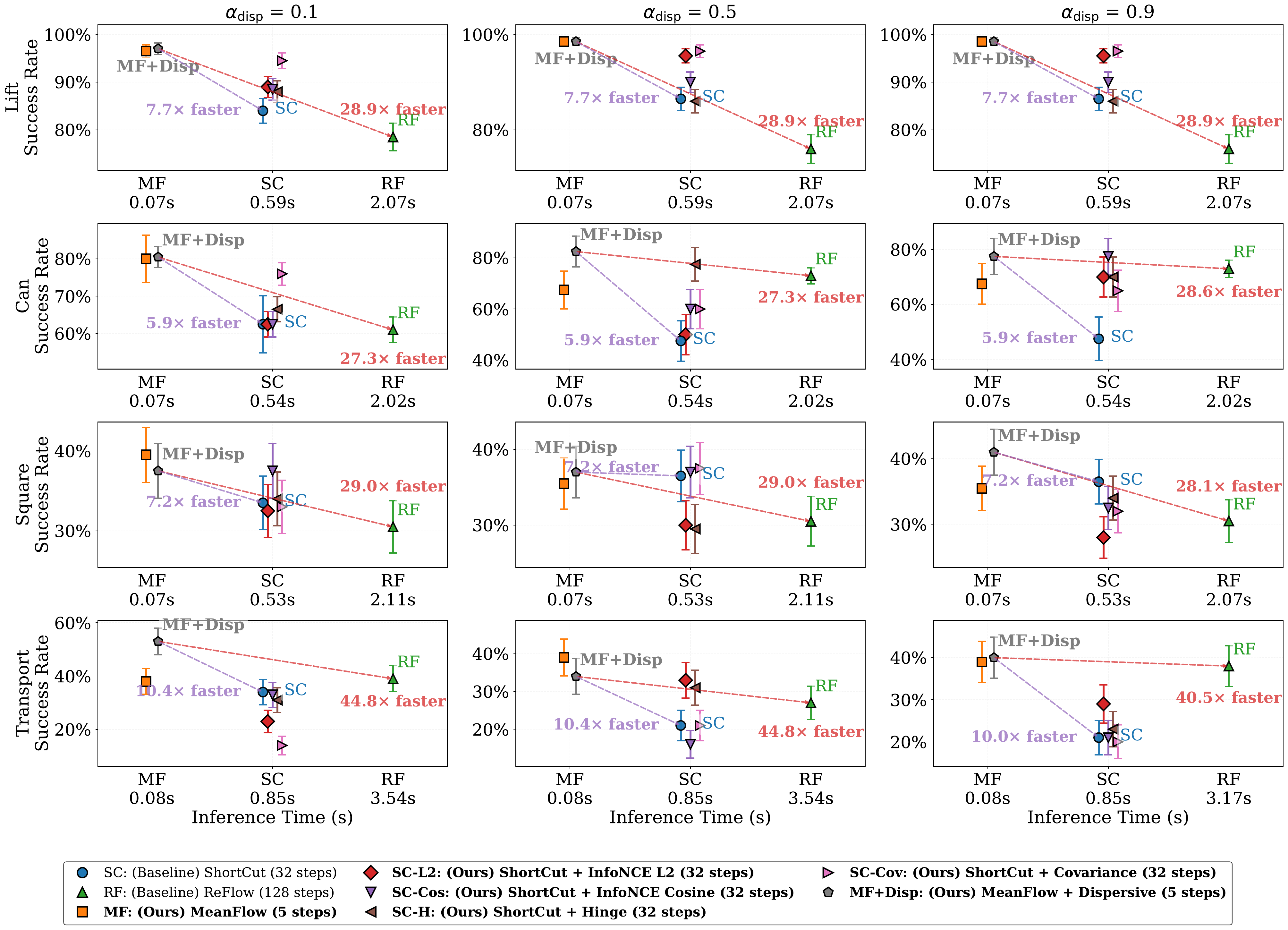}
\caption{Success rate vs. inference frequency trade-off across different weighting factors ($\alpha_{\text{disp}} = 0.1, 0.5, 0.9$) and four tasks. Each point represents a method's performance, with position indicating the trade-off between task success (y-axis) and computational efficiency (x-axis, left = faster). Our methods achieve consistent improvements in both dimensions compared to baselines (ShortCut, ReFlow).}
\label{fig:success_vs_frequency}
\end{figure*}

Task-Specific Performance Patterns: Performance varies significantly across tasks. For \textit{Lift}, MeanFlow achieves peak performance at 5 steps while baselines require 32+ steps. The \textit{Can} task reveals more dramatic differences: baseline methods fail almost entirely at low step counts (1--4 steps), while MeanFlow reaches 60--70\% success at 5 steps, with MF+Disp achieving 75--80\% at $\alpha_{\text{disp}}=0.5$ and $0.9$. Most notably, \textit{Transport} sees baseline methods fail completely at low steps, while MeanFlow achieves 40--50\% success at 5 steps, with MF+Disp reaching 60\% peak performance. MF+Disp consistently outperforms vanilla MF across all tasks, with improvement most pronounced for Transport and Can where dispersive regularization prevents representation collapse.

\subsubsection{\textbf{Efficiency-Performance Trade-off Analysis (RQ1)}}

Figure \ref{fig:success_vs_frequency} analyzes performance-efficiency trade-offs across all methods and task configurations, directly addressing RQ1. The scatter plot compares success rates (y-axis) against inference time (x-axis), where upper-left points represent optimal performance and speed. Inference time is measured as total wall-clock duration across 50 parallel environments.

Optimal Trade-off Achievement: MF+Disp consistently positions in the upper-left region across all tasks and weight configurations, achieving superior success rates and faster inference. Baseline methods (ReFlow at 2--3.5s, ShortCut at 0.6--0.8s) require substantially longer inference, while MeanFlow variants complete inference in 0.07--0.09s, representing a 20--40$\times$ speedup.

\begin{table*}[t]
\caption{Reward Comparison under 1-Step and 5-Step Denoising across Different Dispersive Weights and Tasks}
\label{tab:comprehensive_combined}
\footnotesize
\begin{tabular}{llcc|cccccc|cc|cccccc}
\toprule
& & \multicolumn{8}{c|}{\textbf{1-Step Denoising}} & \multicolumn{8}{c}{\textbf{5-Step Denoising}} \\
& & \multicolumn{2}{c|}{\textit{Baseline}} & \multicolumn{6}{c|}{\cellcolor{lightblue}\textit{Ours}} & \multicolumn{2}{c|}{\textit{Baseline}} & \multicolumn{6}{c}{\cellcolor{lightblue}\textit{Ours}} \\
$\boldsymbol{\alpha_{\text{disp}}}$ & \textbf{Task} & \textbf{RF} & \textbf{SC} & \cellcolor{lightblue}\textbf{SC-L2} & \cellcolor{lightblue}\textbf{SC-Cos} & \cellcolor{lightblue}\textbf{SC-H} & \cellcolor{lightblue}\textbf{SC-Cov} & \cellcolor{lightblue}\textbf{MF} & \cellcolor{lightblue}\textbf{MF+Disp} & \textbf{RF} & \textbf{SC} & \cellcolor{lightblue}\textbf{SC-L2} & \cellcolor{lightblue}\textbf{SC-Cos} & \cellcolor{lightblue}\textbf{SC-H} & \cellcolor{lightblue}\textbf{SC-Cov} & \cellcolor{lightblue}\textbf{MF} & \cellcolor{lightblue}\textbf{MF+Disp} \\
\midrule
\multirow{4}{*}{0.1} & Lift & 40.1 & 43.5 & \cellcolor{lightblue}35.9 & \cellcolor{lightblue}31.4 & \cellcolor{lightblue}86.9 & \cellcolor{lightblue}60.8 & \cellcolor{lightblue}68.9 & \cellcolor{lightblue}\textbf{86.5} & 25.8 & 39.3 & \cellcolor{lightblue}40.9 & \cellcolor{lightblue}36.2 & \cellcolor{lightblue}35.3 & \cellcolor{lightblue}68.2 & \cellcolor{lightblue}78.3 & \cellcolor{lightblue}\textbf{111.2} \\
& Can & 68.0 & 78.9 & \cellcolor{lightblue}68.1 & \cellcolor{lightblue}68.1 & \cellcolor{lightblue}80.3 & \cellcolor{lightblue}65.0 & \cellcolor{lightblue}95.5 & \cellcolor{lightblue}\textbf{98.0} & 1.9 & 70.2 & \cellcolor{lightblue}81.6 & \cellcolor{lightblue}81.6 & \cellcolor{lightblue}90.6 & \cellcolor{lightblue}84.8 & \cellcolor{lightblue}99.5 & \cellcolor{lightblue}\textbf{117.9} \\
& Square & 31.0 & 23.4 & \cellcolor{lightblue}11.0 & \cellcolor{lightblue}21.1 & \cellcolor{lightblue}15.9 & \cellcolor{lightblue}19.1 & \cellcolor{lightblue}23.4 & \cellcolor{lightblue}\textbf{38.0} & 0.0 & 41.0 & \cellcolor{lightblue}16.7 & \cellcolor{lightblue}38.1 & \cellcolor{lightblue}32.1 & \cellcolor{lightblue}32.9 & \cellcolor{lightblue}\textbf{41.1} & \cellcolor{lightblue}38.5 \\
& Transport & 77.6 & 90.3 & \cellcolor{lightblue}63.8 & \cellcolor{lightblue}37.9 & \cellcolor{lightblue}15.9 & \cellcolor{lightblue}8.7 & \cellcolor{lightblue}\textbf{119.9} & \cellcolor{lightblue}87.7 & 0.0 & 78.9 & \cellcolor{lightblue}42.1 & \cellcolor{lightblue}98.9 & \cellcolor{lightblue}32.1 & \cellcolor{lightblue}0.0 & \cellcolor{lightblue}123.4 & \cellcolor{lightblue}\textbf{171.6} \\
\midrule
\multirow{4}{*}{0.5} & Lift & 66.8 & 26.1 & \cellcolor{lightblue}44.1 & \cellcolor{lightblue}60.6 & \cellcolor{lightblue}63.8 & \cellcolor{lightblue}\textbf{81.8} & \cellcolor{lightblue}76.8 & \cellcolor{lightblue}70.4 & 12.0 & 27.4 & \cellcolor{lightblue}75.0 & \cellcolor{lightblue}43.7 & \cellcolor{lightblue}74.4 & \cellcolor{lightblue}68.3 & \cellcolor{lightblue}\textbf{90.2} & \cellcolor{lightblue}75.6 \\
& Can & 76.2 & 84.5 & \cellcolor{lightblue}48.1 & \cellcolor{lightblue}66.0 & \cellcolor{lightblue}81.4 & \cellcolor{lightblue}54.2 & \cellcolor{lightblue}\textbf{107.8} & \cellcolor{lightblue}100.3 & 1.8 & 106.2 & \cellcolor{lightblue}76.3 & \cellcolor{lightblue}104.8 & \cellcolor{lightblue}93.1 & \cellcolor{lightblue}73.3 & \cellcolor{lightblue}96.7 & \cellcolor{lightblue}\textbf{116.3} \\
& Square & 24.9 & 30.2 & \cellcolor{lightblue}12.1 & \cellcolor{lightblue}22.6 & \cellcolor{lightblue}14.6 & \cellcolor{lightblue}16.0 & \cellcolor{lightblue}\textbf{36.8} & \cellcolor{lightblue}30.9 & 1.9 & 37.9 & \cellcolor{lightblue}23.0 & \cellcolor{lightblue}\textbf{46.0} & \cellcolor{lightblue}34.4 & \cellcolor{lightblue}33.6 & \cellcolor{lightblue}38.7 & \cellcolor{lightblue}41.7 \\
& Transport & 26.6 & 68.1 & \cellcolor{lightblue}82.5 & \cellcolor{lightblue}14.5 & \cellcolor{lightblue}39.1 & \cellcolor{lightblue}12.8 & \cellcolor{lightblue}\textbf{114.7} & \cellcolor{lightblue}99.6 & 29.2 & 62.6 & \cellcolor{lightblue}0.0 & \cellcolor{lightblue}59.1 & \cellcolor{lightblue}85.8 & \cellcolor{lightblue}22.9 & \cellcolor{lightblue}\textbf{122.3} & \cellcolor{lightblue}106.1 \\
\midrule
\multirow{4}{*}{0.9} & Lift & 66.8 & 68.1 & \cellcolor{lightblue}39.8 & \cellcolor{lightblue}27.6 & \cellcolor{lightblue}56.7 & \cellcolor{lightblue}43.2 & \cellcolor{lightblue}\textbf{76.8} & \cellcolor{lightblue}68.7 & 12.0 & 62.6 & \cellcolor{lightblue}\textbf{116.0} & \cellcolor{lightblue}30.9 & \cellcolor{lightblue}67.5 & \cellcolor{lightblue}161.5 & \cellcolor{lightblue}90.2 & \cellcolor{lightblue}76.5 \\
& Can & 76.2 & 84.5 & \cellcolor{lightblue}57.6 & \cellcolor{lightblue}77.9 & \cellcolor{lightblue}67.0 & \cellcolor{lightblue}88.3 & \cellcolor{lightblue}107.8 & \cellcolor{lightblue}\textbf{104.8} & 1.8 & 106.2 & \cellcolor{lightblue}98.5 & \cellcolor{lightblue}89.2 & \cellcolor{lightblue}79.8 & \cellcolor{lightblue}67.2 & \cellcolor{lightblue}96.7 & \cellcolor{lightblue}\textbf{112.2} \\
& Square & 24.9 & 30.2 & \cellcolor{lightblue}10.5 & \cellcolor{lightblue}27.0 & \cellcolor{lightblue}13.7 & \cellcolor{lightblue}8.5 & \cellcolor{lightblue}36.8 & \cellcolor{lightblue}\textbf{43.8} & 1.9 & 37.9 & \cellcolor{lightblue}27.1 & \cellcolor{lightblue}35.7 & \cellcolor{lightblue}26.0 & \cellcolor{lightblue}20.4 & \cellcolor{lightblue}38.7 & \cellcolor{lightblue}\textbf{46.2} \\
& Transport & 82.5 & 68.1 & \cellcolor{lightblue}27.3 & \cellcolor{lightblue}11.0 & \cellcolor{lightblue}39.8 & \cellcolor{lightblue}1.1 & \cellcolor{lightblue}114.7 & \cellcolor{lightblue}\textbf{131.3} & 0.0 & 62.6 & \cellcolor{lightblue}54.2 & \cellcolor{lightblue}31.2 & \cellcolor{lightblue}67.1 & \cellcolor{lightblue}3.5 & \cellcolor{lightblue}122.3 & \cellcolor{lightblue}\textbf{126.8} \\
\bottomrule
\end{tabular}
\end{table*}

\begin{table}[t]
\caption{Success Rate and Reward under Optimal Step Configurations}
\label{tab:optimal_steps}
\footnotesize
\begin{tabular}{llcc|c|cc|c}
\toprule
& & \multicolumn{3}{c|}{\textbf{Success (\%)}} & \multicolumn{3}{c}{\textbf{Reward}} \\
& & \multicolumn{2}{c|}{\textit{Baseline}} & \cellcolor{lightblue}\textit{Ours} & \multicolumn{2}{c|}{\textit{Baseline}} & \cellcolor{lightblue}\textit{Ours} \\
$\boldsymbol{\alpha_{\text{disp}}}$ & \textbf{Task} & \textbf{RF} & \textbf{SC} & \cellcolor{lightblue}\textbf{MF+Disp} & \textbf{RF} & \textbf{SC} & \cellcolor{lightblue}\textbf{MF+Disp} \\
\midrule
\multirow{4}{*}{0.1} & Lift & 78.5\% & 84.0\% & \cellcolor{lightblue}\textbf{97.0\%} & 56.5 & 38.0 & \cellcolor{lightblue}\textbf{111.2} \\
& Can & 61.0\% & 62.5\% & \cellcolor{lightblue}\textbf{80.5\%} & 86.9 & 89.8 & \cellcolor{lightblue}\textbf{117.9} \\
& Square & 30.5\% & 33.5\% & \cellcolor{lightblue}\textbf{37.5\%} & 30.7 & 34.1 & \cellcolor{lightblue}\textbf{38.5} \\
& Trans. & 39.0\% & 34.0\% & \cellcolor{lightblue}\textbf{53.0\%} & 111.8 & 103.8 & \cellcolor{lightblue}\textbf{171.6} \\
\midrule
\multirow{4}{*}{0.5} & Lift & 76.0\% & 86.5\% & \cellcolor{lightblue}\textbf{98.5\%} & 62.9 & 25.7 & \cellcolor{lightblue}\textbf{75.6} \\
& Can & 73.0\% & 47.5\% & \cellcolor{lightblue}\textbf{82.5\%} & 94.1 & 74.2 & \cellcolor{lightblue}\textbf{116.3} \\
& Square & 30.5\% & 36.5\% & \cellcolor{lightblue}\textbf{37.0\%} & 27.3 & 39.8 & \cellcolor{lightblue}\textbf{41.7} \\
& Trans. & 27.0\% & 21.0\% & \cellcolor{lightblue}\textbf{34.0\%} & 77.2 & 62.0 & \cellcolor{lightblue}\textbf{106.1} \\
\midrule
\multirow{4}{*}{0.9} & Lift & 76.0\% & 86.5\% & \cellcolor{lightblue}\textbf{96.0\%} & 62.9 & 25.7 & \cellcolor{lightblue}\textbf{76.5} \\
& Can & 73.0\% & 47.5\% & \cellcolor{lightblue}\textbf{77.5\%} & 94.1 & 74.2 & \cellcolor{lightblue}\textbf{112.2} \\
& Square & 30.5\% & 36.5\% & \cellcolor{lightblue}\textbf{41.0\%} & 27.3 & 39.8 & \cellcolor{lightblue}\textbf{46.2} \\
& Trans. & 38.0\% & 21.0\% & \cellcolor{lightblue}\textbf{40.0\%} & 115.9 & 62.0 & \cellcolor{lightblue}\textbf{126.8} \\
\bottomrule
\end{tabular}
\end{table}

Task-Specific Analysis: For \textit{Lift}, all methods reach near 100\% success, but baselines require \textcolor{highlight}{20--40$\times$ longer} inference without gains. On \textit{Can}, MF+Disp achieves 75--80\% success in \texttt{0.07s} versus 60--70\% for baselines in \texttt{2--3s} ($\uparrow$\textcolor{highlight}{15--20\%}). For \textit{Transport}, MF+Disp achieves 40--50\% success in \texttt{0.07s} versus 20--25\% for baselines in \texttt{2--3s}, a \textcolor{highlight}{40$\times$ speedup} with \textcolor{highlight}{2$\times$ performance gain}.

Dispersive Regularization Value Demonstration: The comparison between MF and MF+Disp reveals targeted benefits of dispersive regularization. While both achieve similar efficiency, MF+Disp consistently occupies higher positions on the y-axis, particularly for complex tasks. For simple tasks like Lift, improvement is marginal, but for Can, Square, and Transport, dispersive regularization provides $\uparrow$\textcolor{highlight}{10--20 percentage point gains}.

\subsubsection{\textbf{Comprehensive Reward Analysis (RQ2 \& RQ3)}}

Table \ref{tab:comprehensive_combined} presents a detailed analysis of reward values across three key weight configurations ($\alpha_{\text{disp}} \in \{0.1, 0.5, 0.9\}$) and different denoising step counts (1-step vs. 5-step), comparing baseline methods (RF, SC) with our proposed dispersive regularization variants (SC-L2, SC-Cos, SC-H, SC-Cov, MF, MF+Disp). This analysis addresses RQ2 (dispersive regularization effectiveness) and RQ3 (comparison of different variants).

Consistent Superior Performance: MF+Disp achieves the highest reward values in most task-weight configurations, demonstrating the effectiveness of combining MeanFlow with dispersive regularization. Notably, for the Can task at $\alpha_{\text{disp}} = 0.5$ with 1-step denoising, MF+Disp reaches \textbf{100.3} reward, significantly outperforming baselines (RF: 76.2 $\rightarrow$ $\uparrow$\textcolor{highlight}{24.1}, SC: 84.5 $\rightarrow$ $\uparrow$\textcolor{highlight}{15.8}). Similarly, for the Transport task at $\alpha_{\text{disp}} = 0.9$ with 5-step denoising, MF+Disp achieves \textbf{126.8} reward, substantially higher than baseline methods.

Weight Configuration Impact Analysis: The dispersive regularization weight significantly influences performance across different tasks. For the Transport task (the most complex manipulation scenario), $\alpha_{\text{disp}} = 0.1$ achieves exceptional performance with MF+Disp reaching 131.3 and 171.6 for 1-step and 5-step denoising respectively. This suggests that lighter regularization allows the model to capture complex multimodal behaviors required for sophisticated multi-stage manipulation sequences, while heavier weights may over-constrain the representation learning.

Step-wise Performance Comparison: The comparison between 1-step and 5-step denoising reveals that dispersive regularization provides particularly significant improvements in the 1-step setting. For example, in the Can task ($\alpha_{\text{disp}} = 0.1$), MF+Disp shows a 20\% improvement from 1-step (98.0) to 5-step (117.9), while maintaining stability across different configurations. This demonstrates that dispersive regularization is crucial for stabilizing performance when reducing computational steps.

Baseline Method Instability: ReFlow exhibits significant instability across different configurations, with several instances of complete failure (0.0 rewards) in the 5-step denoising scenario. This highlights the robustness advantage of our MeanFlow-based approach, which maintains consistent performance across all tested configurations.

\begin{table*}[t]
\centering
\caption{Per-stage latency breakdown (ms) for Lift task on physical robot, excluding \textbf{Execution}. MF: MeanFlow (Ours), SC: ShortCut, RF: ReFlow. Numbers in parentheses indicate denoising steps. T: Total latency. }
\small
\setlength{\tabcolsep}{4pt}
\begin{tabular}{lcccccccccccccc}
\toprule
\textbf{Planner} & \textbf{Camera} & \textbf{State} & \textbf{Prep.} & \cellcolor{lightblue}\textbf{MF(1)} & \cellcolor{lightblue}\textbf{MF(5)} & \textbf{SC(32)} & \textbf{RF(128)} & \cellcolor{pink!50}\textbf{Plan.} & \textbf{Send} & \cellcolor{lightblue}\textbf{T-MF(1)} & \cellcolor{lightblue}\textbf{T-MF(5)} & \textbf{T-SC(32)} & \textbf{T-RF(128)} \\
\midrule
\textbf{Cartesian}   & 5.4 & 0.1 & 0.4 & \cellcolor{lightblue}2.4 & \cellcolor{lightblue}10.5 & 76.5 & 305.3 & \cellcolor{pink!50}1.7   & 1.1 & \cellcolor{lightblue}11.1 & \cellcolor{lightblue}19.2 & 85.2  & 314.1 \\
\textbf{BiT-RRT*}     & 7.6 & 0.2 & 0.5 & \cellcolor{lightblue}2.4 & \cellcolor{lightblue}11.4 & 77.0 & 306.9 & \cellcolor{pink!50}89.9  & 2.5 & \cellcolor{lightblue}103.1 & \cellcolor{lightblue}112.1 & 177.7 & 407.6 \\
\textbf{RRTConnect*} & 8.4 & 0.2 & 0.4 & \cellcolor{lightblue}2.5 & \cellcolor{lightblue}13.8 & 79.1 & 312.8 & \cellcolor{pink!50}152.4 & 2.8 & \cellcolor{lightblue}166.7 & \cellcolor{lightblue}178.0 & 243.3 & 477.0 \\
\textbf{RRT*}    & 8.7 & 0.2 & 0.5 & \cellcolor{lightblue}2.4 & \cellcolor{lightblue}12.2 & 78.0 & 308.6 & \cellcolor{pink!50}604.8 & 2.6 & \cellcolor{lightblue}619.3 & \cellcolor{lightblue}629.0 & 694.8 & 925.4 \\
\bottomrule
\end{tabular}
\label{tab:latency-no-exec}
\end{table*}

\begin{figure*}[t]
\centering
\includegraphics[width=0.95\textwidth]{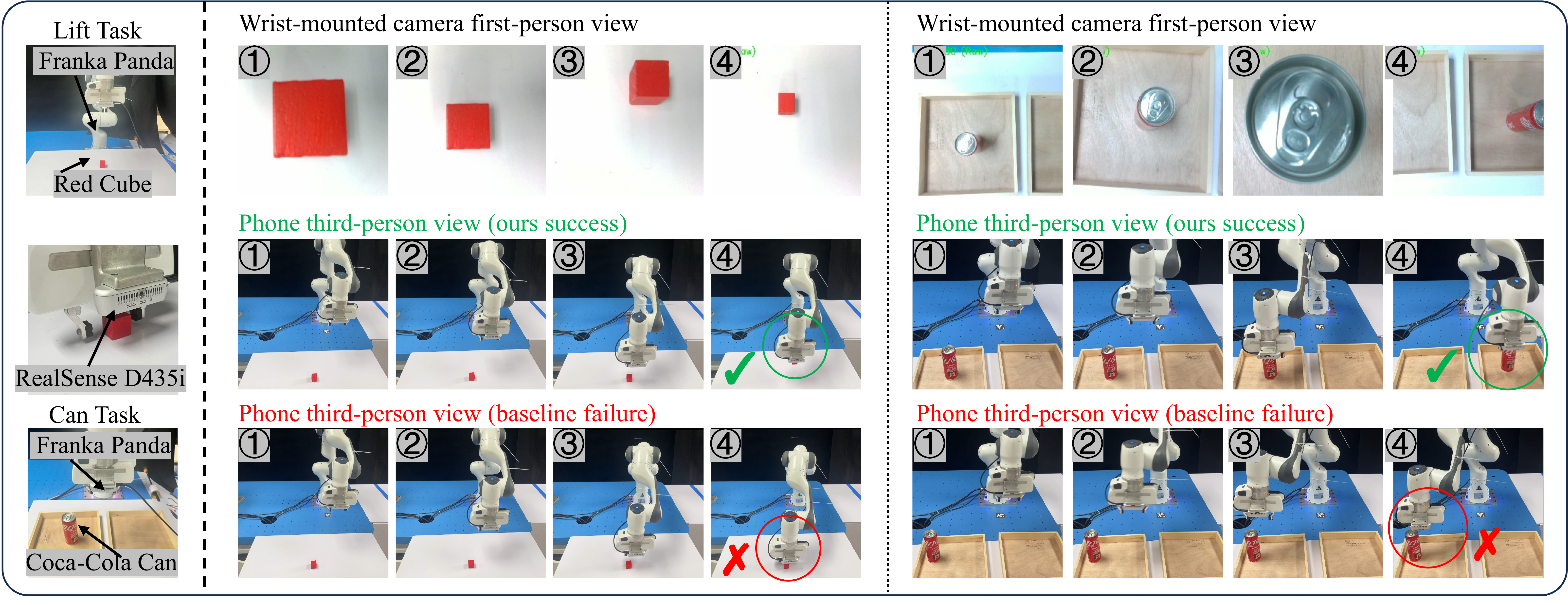}
\caption{
    Real-world deployment on a Franka-Emika-Panda robot for \textbf{Lift} (red cube) and \textbf{Can} (Coca-Cola can) tasks.
    Each task shows wrist-mounted first-person views (top), successful trials with third-person views (middle, green checkmarks), and failure cases (bottom, red crosses).
    The dual-view visualization enables direct comparison between successful and failed execution.
    }
\label{fig:real_robot}
\end{figure*}

\subsubsection{\textbf{Optimal Configuration Analysis (RQ1 \& RQ2)}}

Table \ref{tab:optimal_steps} compares performance under optimal step configurations: MF+Disp (5 steps), ShortCut (32 steps), and ReFlow (128 steps), providing further evidence for RQ1 and RQ2. MF+Disp consistently outperforms baselines in both success rates and rewards across all tasks and dispersive weights. Notably, with only 5 denoising steps, MF+Disp achieves 6.4$\times$ and 25.6$\times$ computational speedup over ShortCut and ReFlow respectively, while maintaining superior task performance, demonstrating that efficiency gains do not compromise solution quality.

\subsection{Physical Experiments (RQ4)}

To validate the practical applicability of DM1 beyond simulation (RQ4), we conduct real-world experiments on a Franka-Emika-Panda robot. The physical deployment tests DM1's ability to transfer from simulation to physical hardware and demonstrates its real-time control capabilities under realistic computational constraints.

\textbf{Experimental Setup:} We deploy DM1 on a 7-Degrees-of-Freedom (DOF) Franka-Emika-Panda robot with an eye-in-hand RGB camera (96$\times$96$\times$3) using an NVIDIA RTX 2080 GPU. We evaluate MeanFlow (1-step and 5-step) against ShortCut (32 steps) and ReFlow (128 steps) baselines. The network receives multimodal observations including 9-dimensional proprioceptive state and wrist camera images, outputting action sequences with horizon $H=4$. We test four motion planning strategies with varying complexity: Cartesian (fastest), BiT-RRT*, RRTConnect*, and RRT* on the Lift task.

\textbf{Latency Analysis:} Table~\ref{tab:latency-no-exec} presents per-stage latency breakdown for the Lift task. Neural network inference time scales dramatically with denoising steps: MeanFlow with 5 steps requires only 10.5-13.8ms compared to 76.5-79.1ms for ShortCut and 305.3-312.8ms for ReFlow (7-29$\times$ speedup). Motion planning complexity creates a substantial bottleneck: Cartesian planning adds only 1.7ms while RRT* requires 604.8ms. Total latency for MeanFlow with 5 steps ranges from 19.2ms (Cartesian) to 629.0ms (RRT*), with variation driven primarily by planning rather than inference.

\textbf{Deployment Insights:} Figure~\ref{fig:real_robot} visualizes our real-world deployment on a Franka-Emika-Panda robot across two representative tasks: Lift and Can. The visualization employs a dual-perspective approach. The top row shows the wrist-mounted camera view that provides the actual visual input to the policy, while the middle and bottom rows present third-person phone camera views that capture the overall manipulation process. For each task, the middle row displays successful executions of our MeanFlow method marked with green checkmarks, whereas the bottom row illustrates failure cases of the ShortCut Flow baseline marked with red crosses.

The results demonstrate that combining MeanFlow's efficient 5-step generation with Cartesian planning achieves optimal balance: 19.2ms total latency enables 50Hz+ control frequency while maintaining smooth, natural motions. In contrast, more complex planning strategies (BiT-RRT*, RRTConnect*, RRT*) incur 112-629ms latency even with efficient MeanFlow inference, limiting control frequencies to 1.6-9Hz. The computational efficiency advantages observed in simulation translate directly to practical real-time control capability, validating DM1's applicability for closed-loop robotic manipulation in real-world environments.

%%% The following command should be issued somewhere in the first column 
%%% of the final page of your paper.
\balance

%%%%%%%%%%%%%%%%%%%%%%%%%%%%%%%%%%%%%%%%%%%%%%%%%%%%%%%%%%%%%%%%%%%%%%%%

\section{Conclusion}

We presented DM1, the first systematic integration of dispersive regularization with MeanFlow for vision-based robotic manipulation, addressing two fundamental challenges in flow-based control: achieving one-step generation efficiency while preventing representation collapse. Our key insight is that applying dispersive regularization to intermediate representations (without architectural modifications) effectively preserves feature diversity while maintaining computational efficiency. Comprehensive experiments on RoboMimic benchmarks and physical Franka-Emika-Panda robot deployment validate DM1's effectiveness. With only 5 denoising steps, DM1 achieves 6.4-25.6$\times$ speedup over baselines (ShortCut 32 steps, ReFlow 128 steps) while improving success rates by 10-20 percentage points on complex tasks. Physical experiments demonstrate 19.2ms total latency enabling 50Hz+ real-time control, validating the approach's practical viability. DM1 demonstrates that efficient one-step generation and robust multimodal representations are not mutually exclusive, a crucial insight for deploying generative models in real-time robotic systems where both speed and expressiveness are critical.

%%%%%%%%%%%%%%%%%%%%%%%%%%%%%%%%%%%%%%%%%%%%%%%%%%%%%%%%%%%%%%%%%%%%%%%%

%%% The acknowledgments section is defined using the "acks" environment
%%% (rather than an unnumbered section). The use of this environment 
%%% ensures the proper identification of the section in the article 
%%% metadata as well as the consistent spelling of the heading.

%%%%%%%%%%%%%%%%%%%%%%%%%%%%%%%%%%%%%%%%%%%%%%%%%%%%%%%%%%%%%%%%%%%%%%%%

%%% The next two lines define, first, the bibliography style to be 
%%% applied, and, second, the bibliography file to be used.

\bibliographystyle{ACM-Reference-Format}
\bibliography{sample}

%%%%%%%%%%%%%%%%%%%%%%%%%%%%%%%%%%%%%%%%%%%%%%%%%%%%%%%%%%%%%%%%%%%%%%%%

\onecolumn
\appendix

\section{Discussion and Analysis}

\subsection{Key Insights and Theoretical Implications}

Our experimental findings provide several crucial insights into simultaneously addressing the two fundamental challenges in flow-based robotic control: one-step generation efficiency and representation collapse.

\textbf{One-Step Generation Insights:} The successful adaptation of MeanFlow to robotic control demonstrates that average velocity-based flow matching can maintain high generation quality while achieving dramatic computational efficiency improvements. The 5-step sampling process proves sufficient for capturing the complexity inherent in sophisticated manipulation tasks, suggesting that true one-step generation (or near one-step with minimal steps) is achievable without sacrificing expressiveness. This finding challenges the conventional wisdom that multi-step inference is necessary for high-quality robotic control.

\textbf{Representation Collapse Insights:} The analysis of dispersive regularization variants reveals important insights into preventing representation collapse in flow-based models. The superior performance of InfoNCE-cosine variants compared to L2-based alternatives suggests that angular diversity in high-dimensional feature spaces is more beneficial than magnitude-based separation for addressing collapse.

\subsection{Limitations and Scope Considerations}

Despite the promising results demonstrated by DM1, several important limitations warrant consideration. The observed performance gains exhibit task-dependent variation, with smaller improvements on relatively simple manipulation tasks like object lifting compared to more complex scenarios involving precision insertion or multi-step transport operations. This pattern suggests that the benefits of our approach are most pronounced when dealing with tasks that require sophisticated behavioral representations, indicating potential areas for further improvement in basic manipulation skills.

The dispersive regularization mechanism introduces additional hyperparameter complexity, requiring careful tuning of weight coefficients and temperature parameters to achieve optimal performance. This sensitivity may limit the method's applicability in scenarios where extensive hyperparameter optimization is not feasible, suggesting the need for more robust or adaptive regularization strategies in future work.

Our experimental evaluation focuses primarily on tabletop manipulation scenarios within controlled laboratory environments. The generalization of these findings to more diverse robotic applications, including mobile manipulation, dexterous hand control, and multi-robot coordination scenarios, remains an open question that would benefit from additional empirical investigation. Furthermore, the long-term stability and adaptation capabilities of the proposed method under varying environmental conditions and task distributions require further study.

\subsection{Future Research Directions}

The success of DM1 opens several promising avenues for future research and development. The development of adaptive mechanisms for automatically adjusting dispersive regularization strength during training could significantly improve the robustness and applicability of the approach while reducing the burden of manual hyperparameter tuning. Such adaptive methods might leverage meta-learning principles or online optimization techniques to dynamically balance representation diversity with task-specific performance objectives.

The integration of additional sensory modalities, particularly tactile feedback and force sensing, represents a natural extension of the current framework that could substantially enhance performance on contact-rich manipulation tasks. The probabilistic nature of flow-based models makes them well-suited for incorporating multi-modal sensor information, potentially leading to more robust and capable robotic control systems.

The hierarchical extension of DM1 to incorporate higher-level task planning and goal specification could enable more complex, long-horizon robotic behaviors while maintaining the computational efficiency advantages of the underlying MeanFlow formulation. Such extensions might involve multi-scale temporal modeling or the integration of symbolic planning components with the continuous control framework established in this work.

\section{MeanFlow Mathematical Derivation for Robotic Control}

This appendix provides detailed mathematical derivations that complement the main text, focusing on step-by-step proofs and robotic-specific implementation details omitted in Section 3 for brevity.

\subsection{MeanFlow Identity and Total Derivative}

Here we provide the detailed derivation of the MeanFlow identity presented in Equation (7) of the main text. Starting from the displacement identity $(t-r)u(z_t,r,t) = z_t - z_r$, we take the derivative with respect to $t$:

\begin{align}
\frac{d}{dt}[(t-r)u(z_t, r, t)] &= \frac{d}{dt}\int_r^t v_\tau(z_\tau, \tau) d\tau\\
&= v_t(z_t, t) \quad \text{(by fundamental theorem of calculus)}
\end{align}

Applying the product rule to the left side:

\begin{align}
\frac{d}{dt}[(t-r)u(z_t, r, t)] &= u(z_t, r, t) + (t-r)\frac{d}{dt}u(z_t, r, t)
\end{align}

Equating these two expressions and rearranging:

\begin{equation}
u(z_t, r, t) + (t-r)\frac{d}{dt}u(z_t, r, t) = v_t(z_t, t)
\label{eq:meanflow_intermediate}
\end{equation}

which yields the MeanFlow identity:

\begin{equation}
\boxed{u(z_t, r, t) = v_t(z_t, t) - (t-r)\frac{d}{dt}u(z_t, r, t)}
\label{eq:meanflow_identity_appendix}
\end{equation}

The total derivative in Eq.~\eqref{eq:meanflow_intermediate} requires careful computation using the multivariate chain rule. Since $u(z_t, r, t)$ depends on $t$ both explicitly and through $z_t(t)$, we apply:

\begin{align}
\frac{d}{dt}u(z_t, r, t) &= \frac{dz_t}{dt} \cdot \nabla_{z_t} u + \frac{dr}{dt} \frac{\partial u}{\partial r} + \frac{dt}{dt} \frac{\partial u}{\partial t}
\end{align}

Since $r$ is held constant during the time evolution, $\frac{dr}{dt} = 0$, and $\frac{dt}{dt} = 1$:

\begin{align}
\frac{d}{dt}u(z_t, r, t) &= v_t(z_t, t) \cdot \nabla_{z_t} u + \frac{\partial u}{\partial t}
\end{align}

where $v_t(z_t, t) = a - \epsilon$ follows the MeanFlow training objective introduced in the main text (with $a$ the target action and $\epsilon$ the Gaussian reference noise) and represents the instantaneous velocity. This can be efficiently computed using forward-mode automatic differentiation with Jacobian-Vector Products (JVPs) using the tangent vector $(v_t(z_t, t), 0, 1)$ with respect to variables $(z_t, r, t)$.

\subsection{Temporal Sampling and Inference Strategies}
\label{subsec:temporal_sampling}

Unlike computer vision applications where uniform sampling often suffices, robotic control requires careful temporal sampling to ensure stable learning and smooth action generation. We employ a logit-normal distribution for sampling $(t,r)$ pairs during training:

\begin{align}
\xi &\sim \mathcal{N}(\mu, \sigma^2) \quad \text{where } \mu = -0.4, \sigma = 1.0\\
\tilde{t}, \tilde{r} &= \text{sigmoid}(\xi) = \frac{1}{1 + e^{-\xi}}
\end{align}

We maintain the constraint $r \leq t$ by drawing two independent samples $\tilde{t}_1, \tilde{t}_2$ via the above transformation and then setting $t = \max(\tilde{t}_1, \tilde{t}_2)$ and $r = \min(\tilde{t}_1, \tilde{t}_2)$. This logit-normal sampling concentrates time pairs near trajectory endpoints, encouraging the model to learn consistent start-to-end mappings.

To balance between flow consistency and trajectory diversity, we enforce $r = t$ (instantaneous velocity) for a fraction $\rho$ of training samples (typically $\rho = 0.5$ for robotic tasks), ensuring the model learns both average and instantaneous velocity fields.

During robot execution, MeanFlow enables efficient one-step generation by directly evaluating the average velocity over the full interval $[0,1]$:

\begin{equation}
a_{\text{pred}} = z_0 + u_\theta(z_0, 0, 1, o_t)
\end{equation}

where $z_0 \sim \mathcal{N}(0, I)$ is sampled Gaussian noise. This eliminates the multi-step numerical ODE integration required by traditional flow matching ($K=50$--$100$ steps), reducing inference time from $\sim$100ms to $\sim$5--7ms for typical 7-DOF robot actions, a 15--20$\times$ speedup critical for real-time control at 10--14Hz policy frequency.

\subsection{Key Differences from Computer Vision Applications}

Our MeanFlow adaptation for robotics differs from computer vision implementations in several important ways. Unlike image generation where classifier-free guidance improves quality, robot actions are always conditioned on current observations $o_t$, and unconditional action generation is not meaningful in control contexts. We incorporate hard action bounds $[a_{\min}, a_{\max}]$ through clipping during training and inference to ensure generated actions remain within physically feasible ranges for robot actuators (e.g., joint limits, velocity constraints).

Robot actions must exhibit temporal smoothness for stable physical execution, as abrupt velocity changes can damage hardware or cause tracking errors. Our logit-normal sampling strategy concentrates samples near the trajectory boundaries to encourage coherent action sequences across the planning horizon. The conditioning function must fuse both proprioceptive state $s_t \in \mathbb{R}^{d_s}$ (joint positions, velocities) and high-dimensional visual observations $I_t \in \mathbb{R}^{C \times H \times W}$ through learned encoders, requiring careful architectural design to balance different modalities. Unlike offline image generation, robotic control demands low-latency inference ($<$100ms) to enable reactive closed-loop control, making one-step generation essential.

\subsection{Computational Complexity Analysis}

The computational complexity of MeanFlow training per iteration consists of the forward pass with complexity $O(B \cdot d_a \cdot H)$ for batch size $B$, action dimension $d_a$, and horizon $H$, and the JVP computation with complexity $O(B \cdot d_a \cdot H \cdot N_{\theta})$ where $N_{\theta}$ is the number of network parameters (which dominates for large networks). The total training complexity is $O(B \cdot d_a \cdot H \cdot N_{\theta})$ per iteration, comparable to standard flow matching training.

However, inference complexity differs dramatically. MeanFlow inference requires $O(d_a \cdot H \cdot N_{\theta})$ for a single forward pass, while traditional flow matching requires $O(K \cdot d_a \cdot H \cdot N_{\theta})$ for $K$ forward passes during ODE integration. With $K \sim 50$--$100$ for standard methods, this yields the observed 20--40$\times$ speedup in practice.

\section{Training and Inference Algorithms}
\label{sec:algorithms}

This appendix provides the detailed algorithmic descriptions for DM1 training and inference procedures.

\begin{algorithm}[H]
\caption{DM1 Training with Dispersive Regularization}
\label{alg:dm1_training}
\begin{algorithmic}[1]
\REQUIRE Dataset $\mathcal{D} = \{(s_t, a_t)\}$, dispersive weight $\alpha_{\text{disp}}$, monitored layer index $l$
\ENSURE Trained policy network $\pi_\theta$
\STATE Initialize network parameters $\theta$
\FOR{each epoch}
    \FOR{each batch $\mathcal{B} = \{(s_t^{(i)}, a_t^{(i)})\}_{i=1}^B$ from $\mathcal{D}$}
        \STATE Encode observations: $z^{(i)} = \text{Encoder}_\theta(s_t^{(i)})$
        \STATE Sample Gaussian noise $\epsilon^{(i)} \sim \mathcal{N}(0, I)$ matching the action horizon and dimension
        \STATE Sample $(r^{(i)}, t^{(i)})$ from the logit-normal scheme in Section~\ref{subsec:temporal_sampling}
        \STATE Form interpolants $z_{r}^{(i)} = (1-r^{(i)})\epsilon^{(i)} + r^{(i)}a_t^{(i)}$ and $z_{t}^{(i)} = (1-t^{(i)})\epsilon^{(i)} + t^{(i)}a_t^{(i)}$ (retaining $z_{r}^{(i)}$ for the derivative computation in Eq.~\eqref{eq:meanflow_intermediate})
        \STATE Compute target velocity $v^{\star(i)} = a_t^{(i)} - \epsilon^{(i)}$
        \STATE Predict velocity: $\hat{v}^{(i)} = v_\theta\big(z_{t}^{(i)}, t^{(i)}, z^{(i)}\big)$
        \STATE Extract intermediate representations $\mathbf{H}^{(l)} = \{h^{(l,i)}\}_{i=1}^B$ from layer $l$
        \STATE MeanFlow loss: $\mathcal{L}_{\text{MF}} = \frac{1}{B}\sum_{i=1}^B \|\hat{v}^{(i)} - v^{\star(i)}\|_2^2$
        \STATE Dispersive loss: $\mathcal{L}_{\text{disp}} = \text{DispersiveLoss}(\mathbf{H}^{(l)})$
        \STATE Total loss: $\mathcal{L}_{\text{total}} = \mathcal{L}_{\text{MF}} + \alpha_{\text{disp}} \mathcal{L}_{\text{disp}}$
        \STATE Update parameters: $\theta \leftarrow \theta - \eta \nabla_\theta \mathcal{L}_{\text{total}}$
    \ENDFOR
\ENDFOR
\RETURN $\pi_\theta$
\end{algorithmic}
\end{algorithm}

\begin{algorithm}[H]
\caption{DM1 One-Step Inference}
\label{alg:dm1_inference}
\begin{algorithmic}[1]
\REQUIRE Observation $s_t$, trained policy $\pi_\theta$
\ENSURE Action trajectory $a_t$
\STATE Sample initial noise $z_0 \sim \mathcal{N}(0, I)$
\STATE Encode observation: $z = \text{Encoder}_\theta(s_t)$
\STATE Evaluate average velocity: $\bar{v} = u_\theta(z_0, 0, 1, z)$
\STATE Generate one-step action: $a_t = z_0 + \bar{v}$
\RETURN $a_t$
\end{algorithmic}
\end{algorithm}

\section{Detailed Performance Analysis Across Denoising Steps}

% Comprehensive Success Rates Table (Table 1 format) - Integrating All Weight Configurations
\begin{table*}[htbp]
\centering
\caption{Comprehensive Success Rates Across All Weight Configurations and Key Denoising Steps}
\label{tab:comprehensive_success_rates_detailed}
\scriptsize

% 1-step and 2-step denoising
\begin{tabular}{llcc|cccccc|cc|cccccc}
\toprule
& & \multicolumn{8}{c|}{\textbf{1-Step Denoising}} & \multicolumn{8}{c}{\textbf{2-Step Denoising}} \\
& & \multicolumn{2}{c|}{\textit{Baseline}} & \multicolumn{6}{c|}{\cellcolor{lightblue}\textit{Ours}} & \multicolumn{2}{c|}{\textit{Baseline}} & \multicolumn{6}{c}{\cellcolor{lightblue}\textit{Ours}} \\
$\boldsymbol{\alpha_{\text{disp}}}$ & \textbf{Task} & \textbf{RF} & \textbf{SC} & \cellcolor{lightblue}\textbf{SC-L2} & \cellcolor{lightblue}\textbf{SC-Cos} & \cellcolor{lightblue}\textbf{SC-H} & \cellcolor{lightblue}\textbf{SC-Cov} & \cellcolor{lightblue}\textbf{MF} & \cellcolor{lightblue}\textbf{MF+Disp} & \textbf{RF} & \textbf{SC} & \cellcolor{lightblue}\textbf{SC-L2} & \cellcolor{lightblue}\textbf{SC-Cos} & \cellcolor{lightblue}\textbf{SC-H} & \cellcolor{lightblue}\textbf{SC-Cov} & \cellcolor{lightblue}\textbf{MF} & \cellcolor{lightblue}\textbf{MF+Disp} \\
\midrule
\multirow{4}{*}{0.1} & Lift & 0.625 & 0.890 & \cellcolor{lightblue}0.855 & \cellcolor{lightblue}0.835 & \cellcolor{lightblue}0.860 & \cellcolor{lightblue}0.905 & \cellcolor{lightblue}\textbf{0.935} & \cellcolor{lightblue}\textbf{0.975} & 0.020 & 0.925 & \cellcolor{lightblue}0.865 & \cellcolor{lightblue}0.865 & \cellcolor{lightblue}0.860 & \cellcolor{lightblue}0.905 & \cellcolor{lightblue}\textbf{0.970} & \cellcolor{lightblue}\textbf{0.980} \\
& Can & 0.535 & 0.500 & \cellcolor{lightblue}0.595 & \cellcolor{lightblue}0.595 & \cellcolor{lightblue}0.595 & \cellcolor{lightblue}0.485 & \cellcolor{lightblue}\textbf{0.750} & \cellcolor{lightblue}\textbf{0.740} & 0.000 & 0.625 & \cellcolor{lightblue}0.630 & \cellcolor{lightblue}0.630 & \cellcolor{lightblue}0.660 & \cellcolor{lightblue}0.635 & \cellcolor{lightblue}\textbf{0.825} & \cellcolor{lightblue}0.785 \\
& Square & 0.290 & 0.210 & \cellcolor{lightblue}0.135 & \cellcolor{lightblue}0.205 & \cellcolor{lightblue}0.170 & \cellcolor{lightblue}0.145 & \cellcolor{lightblue}0.235 & \cellcolor{lightblue}\textbf{0.350} & 0.000 & 0.295 & \cellcolor{lightblue}0.180 & \cellcolor{lightblue}0.260 & \cellcolor{lightblue}0.190 & \cellcolor{lightblue}0.180 & \cellcolor{lightblue}\textbf{0.300} & \cellcolor{lightblue}\textbf{0.345} \\
& Transport & 0.270 & 0.250 & \cellcolor{lightblue}0.220 & \cellcolor{lightblue}0.140 & \cellcolor{lightblue}0.170 & \cellcolor{lightblue}0.030 & \cellcolor{lightblue}\textbf{0.390} & \cellcolor{lightblue}0.280 & 0.000 & 0.230 & \cellcolor{lightblue}0.240 & \cellcolor{lightblue}0.210 & \cellcolor{lightblue}0.190 & \cellcolor{lightblue}0.050 & \cellcolor{lightblue}0.290 & \cellcolor{lightblue}\textbf{0.300} \\
\midrule
\multirow{4}{*}{0.5} & Lift & 0.620 & 0.900 & \cellcolor{lightblue}0.750 & \cellcolor{lightblue}0.880 & \cellcolor{lightblue}0.810 & \cellcolor{lightblue}0.880 & \cellcolor{lightblue}\textbf{0.965} & \cellcolor{lightblue}\textbf{0.985} & 0.000 & 0.870 & \cellcolor{lightblue}0.900 & \cellcolor{lightblue}0.965 & \cellcolor{lightblue}0.810 & \cellcolor{lightblue}0.855 & \cellcolor{lightblue}\textbf{0.985} & \cellcolor{lightblue}\textbf{0.985} \\
& Can & 0.550 & 0.625 & \cellcolor{lightblue}0.375 & \cellcolor{lightblue}0.525 & \cellcolor{lightblue}0.575 & \cellcolor{lightblue}0.475 & \cellcolor{lightblue}\textbf{0.800} & \cellcolor{lightblue}0.700 & 0.000 & 0.775 & \cellcolor{lightblue}0.575 & \cellcolor{lightblue}0.650 & \cellcolor{lightblue}0.825 & \cellcolor{lightblue}0.525 & \cellcolor{lightblue}\textbf{0.925} & \cellcolor{lightblue}\textbf{0.800} \\
& Square & 0.235 & 0.300 & \cellcolor{lightblue}0.130 & \cellcolor{lightblue}0.225 & \cellcolor{lightblue}0.140 & \cellcolor{lightblue}0.165 & \cellcolor{lightblue}\textbf{0.340} & \cellcolor{lightblue}\textbf{0.320} & 0.000 & 0.355 & \cellcolor{lightblue}0.195 & \cellcolor{lightblue}0.330 & \cellcolor{lightblue}0.175 & \cellcolor{lightblue}0.205 & \cellcolor{lightblue}0.320 & \cellcolor{lightblue}\textbf{0.365} \\
& Transport & 0.080 & 0.200 & \cellcolor{lightblue}0.270 & \cellcolor{lightblue}0.050 & \cellcolor{lightblue}0.130 & \cellcolor{lightblue}0.060 & \cellcolor{lightblue}\textbf{0.350} & \cellcolor{lightblue}\textbf{0.310} & 0.120 & 0.180 & \cellcolor{lightblue}0.000 & \cellcolor{lightblue}0.080 & \cellcolor{lightblue}0.210 & \cellcolor{lightblue}0.080 & \cellcolor{lightblue}0.320 & \cellcolor{lightblue}\textbf{0.460} \\
\midrule
\multirow{4}{*}{0.9} & Lift & 0.620 & 0.900 & \cellcolor{lightblue}0.680 & \cellcolor{lightblue}0.900 & \cellcolor{lightblue}0.765 & \cellcolor{lightblue}0.870 & \cellcolor{lightblue}\textbf{0.965} & \cellcolor{lightblue}\textbf{0.965} & 0.000 & 0.870 & \cellcolor{lightblue}0.760 & \cellcolor{lightblue}0.915 & \cellcolor{lightblue}0.740 & \cellcolor{lightblue}0.885 & \cellcolor{lightblue}\textbf{0.985} & \cellcolor{lightblue}0.955 \\
& Can & 0.550 & 0.625 & \cellcolor{lightblue}0.425 & \cellcolor{lightblue}0.650 & \cellcolor{lightblue}0.525 & \cellcolor{lightblue}0.675 & \cellcolor{lightblue}\textbf{0.800} & \cellcolor{lightblue}0.700 & 0.000 & 0.775 & \cellcolor{lightblue}0.650 & \cellcolor{lightblue}0.775 & \cellcolor{lightblue}0.600 & \cellcolor{lightblue}0.525 & \cellcolor{lightblue}\textbf{0.925} & \cellcolor{lightblue}0.825 \\
& Square & 0.235 & 0.300 & \cellcolor{lightblue}0.125 & \cellcolor{lightblue}0.305 & \cellcolor{lightblue}0.145 & \cellcolor{lightblue}0.100 & \cellcolor{lightblue}\textbf{0.340} & \cellcolor{lightblue}\textbf{0.395} & 0.000 & 0.355 & \cellcolor{lightblue}0.190 & \cellcolor{lightblue}0.390 & \cellcolor{lightblue}0.220 & \cellcolor{lightblue}0.110 & \cellcolor{lightblue}0.320 & \cellcolor{lightblue}0.360 \\
& Transport & 0.270 & 0.200 & \cellcolor{lightblue}0.100 & \cellcolor{lightblue}0.040 & \cellcolor{lightblue}0.130 & \cellcolor{lightblue}0.010 & \cellcolor{lightblue}\textbf{0.350} & \cellcolor{lightblue}\textbf{0.400} & 0.000 & 0.180 & \cellcolor{lightblue}0.130 & \cellcolor{lightblue}0.090 & \cellcolor{lightblue}0.220 & \cellcolor{lightblue}0.090 & \cellcolor{lightblue}0.320 & \cellcolor{lightblue}0.320 \\
\bottomrule
\end{tabular}

\vspace{0.5em}

% 5-step and 8-step denoising
\begin{tabular}{llcc|cccccc|cc|cccccc}
\toprule
& & \multicolumn{8}{c|}{\textbf{5-Step Denoising}} & \multicolumn{8}{c}{\textbf{8-Step Denoising}} \\
& & \multicolumn{2}{c|}{\textit{Baseline}} & \multicolumn{6}{c|}{\cellcolor{lightblue}\textit{Ours}} & \multicolumn{2}{c|}{\textit{Baseline}} & \multicolumn{6}{c}{\cellcolor{lightblue}\textit{Ours}} \\
$\boldsymbol{\alpha_{\text{disp}}}$ & \textbf{Task} & \textbf{RF} & \textbf{SC} & \cellcolor{lightblue}\textbf{SC-L2} & \cellcolor{lightblue}\textbf{SC-Cos} & \cellcolor{lightblue}\textbf{SC-H} & \cellcolor{lightblue}\textbf{SC-Cov} & \cellcolor{lightblue}\textbf{MF} & \cellcolor{lightblue}\textbf{MF+Disp} & \textbf{RF} & \textbf{SC} & \cellcolor{lightblue}\textbf{SC-L2} & \cellcolor{lightblue}\textbf{SC-Cos} & \cellcolor{lightblue}\textbf{SC-H} & \cellcolor{lightblue}\textbf{SC-Cov} & \cellcolor{lightblue}\textbf{MF} & \cellcolor{lightblue}\textbf{MF+Disp} \\
\midrule
\multirow{4}{*}{0.1} & Lift & 0.305 & 0.890 & \cellcolor{lightblue}0.845 & \cellcolor{lightblue}0.875 & \cellcolor{lightblue}0.795 & \cellcolor{lightblue}0.950 & \cellcolor{lightblue}\textbf{0.965} & \cellcolor{lightblue}\textbf{0.970} & 0.735 & 0.845 & \cellcolor{lightblue}0.865 & \cellcolor{lightblue}0.875 & \cellcolor{lightblue}0.880 & \cellcolor{lightblue}0.940 & \cellcolor{lightblue}\textbf{0.950} & \cellcolor{lightblue}\textbf{0.980} \\
& Can & 0.020 & 0.575 & \cellcolor{lightblue}0.630 & \cellcolor{lightblue}0.630 & \cellcolor{lightblue}0.675 & \cellcolor{lightblue}0.660 & \cellcolor{lightblue}\textbf{0.800} & \cellcolor{lightblue}\textbf{0.805} & 0.205 & 0.475 & \cellcolor{lightblue}\textbf{0.625} & \cellcolor{lightblue}\textbf{0.625} & \cellcolor{lightblue}0.685 & \cellcolor{lightblue}0.615 & \cellcolor{lightblue}\textbf{0.625} & \cellcolor{lightblue}\textbf{0.780} \\
& Square & 0.000 & \textbf{0.370} & \cellcolor{lightblue}0.190 & \cellcolor{lightblue}0.360 & \cellcolor{lightblue}0.290 & \cellcolor{lightblue}0.325 & \cellcolor{lightblue}\textbf{0.395} & \cellcolor{lightblue}\textbf{0.375} & 0.210 & 0.345 & \cellcolor{lightblue}0.215 & \cellcolor{lightblue}0.280 & \cellcolor{lightblue}0.250 & \cellcolor{lightblue}0.310 & \cellcolor{lightblue}0.295 & \cellcolor{lightblue}0.365 \\
& Transport & 0.000 & 0.240 & \cellcolor{lightblue}0.180 & \cellcolor{lightblue}0.340 & \cellcolor{lightblue}0.290 & \cellcolor{lightblue}0.000 & \cellcolor{lightblue}0.380 & \cellcolor{lightblue}\textbf{0.530} & 0.000 & 0.300 & \cellcolor{lightblue}0.230 & \cellcolor{lightblue}0.310 & \cellcolor{lightblue}0.250 & \cellcolor{lightblue}0.000 & \cellcolor{lightblue}0.370 & \cellcolor{lightblue}\textbf{0.430} \\
\midrule
\multirow{4}{*}{0.5} & Lift & 0.125 & 0.905 & \cellcolor{lightblue}0.930 & \cellcolor{lightblue}0.915 & \cellcolor{lightblue}0.820 & \cellcolor{lightblue}0.890 & \cellcolor{lightblue}\textbf{0.985} & \cellcolor{lightblue}\textbf{0.985} & 0.700 & 0.865 & \cellcolor{lightblue}0.940 & \cellcolor{lightblue}0.920 & \cellcolor{lightblue}0.795 & \cellcolor{lightblue}0.925 & \cellcolor{lightblue}\textbf{0.955} & \cellcolor{lightblue}0.965 \\
& Can & 0.015 & \textbf{0.750} & \cellcolor{lightblue}0.575 & \cellcolor{lightblue}0.775 & \cellcolor{lightblue}0.700 & \cellcolor{lightblue}0.575 & \cellcolor{lightblue}0.675 & \cellcolor{lightblue}\textbf{0.825} & 0.265 & 0.500 & \cellcolor{lightblue}0.525 & \cellcolor{lightblue}0.650 & \cellcolor{lightblue}0.500 & \cellcolor{lightblue}0.525 & \cellcolor{lightblue}\textbf{0.625} & \cellcolor{lightblue}0.650 \\
& Square & 0.010 & \textbf{0.375} & \cellcolor{lightblue}0.230 & \cellcolor{lightblue}\textbf{0.385} & \cellcolor{lightblue}0.310 & \cellcolor{lightblue}0.305 & \cellcolor{lightblue}0.355 & \cellcolor{lightblue}\textbf{0.370} & 0.220 & 0.365 & \cellcolor{lightblue}0.275 & \cellcolor{lightblue}0.370 & \cellcolor{lightblue}0.280 & \cellcolor{lightblue}0.305 & \cellcolor{lightblue}0.325 & \cellcolor{lightblue}0.335 \\
& Transport & 0.100 & 0.190 & \cellcolor{lightblue}0.000 & \cellcolor{lightblue}0.190 & \cellcolor{lightblue}0.300 & \cellcolor{lightblue}0.090 & \cellcolor{lightblue}\textbf{0.390} & \cellcolor{lightblue}0.340 & 0.230 & 0.300 & \cellcolor{lightblue}0.010 & \cellcolor{lightblue}0.150 & \cellcolor{lightblue}0.310 & \cellcolor{lightblue}0.040 & \cellcolor{lightblue}\textbf{0.430} & \cellcolor{lightblue}0.380 \\
\midrule
\multirow{4}{*}{0.9} & Lift & 0.125 & 0.905 & \cellcolor{lightblue}0.835 & \cellcolor{lightblue}0.925 & \cellcolor{lightblue}0.795 & \cellcolor{lightblue}0.950 & \cellcolor{lightblue}\textbf{0.985} & \cellcolor{lightblue}\textbf{0.960} & 0.700 & 0.865 & \cellcolor{lightblue}0.905 & \cellcolor{lightblue}0.885 & \cellcolor{lightblue}0.815 & \cellcolor{lightblue}0.955 & \cellcolor{lightblue}\textbf{0.955} & \cellcolor{lightblue}0.965 \\
& Can & 0.015 & \textbf{0.750} & \cellcolor{lightblue}0.675 & \cellcolor{lightblue}\textbf{0.750} & \cellcolor{lightblue}0.625 & \cellcolor{lightblue}0.475 & \cellcolor{lightblue}0.675 & \cellcolor{lightblue}\textbf{0.775} & 0.265 & 0.500 & \cellcolor{lightblue}0.625 & \cellcolor{lightblue}0.575 & \cellcolor{lightblue}0.625 & \cellcolor{lightblue}0.675 & \cellcolor{lightblue}\textbf{0.625} & \cellcolor{lightblue}\textbf{0.750} \\
& Square & 0.010 & \textbf{0.375} & \cellcolor{lightblue}0.270 & \cellcolor{lightblue}0.335 & \cellcolor{lightblue}0.250 & \cellcolor{lightblue}0.210 & \cellcolor{lightblue}0.355 & \cellcolor{lightblue}\textbf{0.410} & 0.220 & 0.365 & \cellcolor{lightblue}0.330 & \cellcolor{lightblue}0.340 & \cellcolor{lightblue}0.280 & \cellcolor{lightblue}0.220 & \cellcolor{lightblue}0.325 & \cellcolor{lightblue}0.340 \\
& Transport & 0.000 & 0.190 & \cellcolor{lightblue}0.200 & \cellcolor{lightblue}0.120 & \cellcolor{lightblue}0.250 & \cellcolor{lightblue}0.020 & \cellcolor{lightblue}\textbf{0.390} & \cellcolor{lightblue}\textbf{0.400} & 0.010 & 0.300 & \cellcolor{lightblue}0.240 & \cellcolor{lightblue}0.120 & \cellcolor{lightblue}0.260 & \cellcolor{lightblue}0.080 & \cellcolor{lightblue}\textbf{0.430} & \cellcolor{lightblue}0.410 \\
\bottomrule
\end{tabular}

\vspace{0.5em}

% 16-step and 32-step denoising
\begin{tabular}{llcc|cccccc|cc|cccccc}
\toprule
& & \multicolumn{8}{c|}{\textbf{16-Step Denoising}} & \multicolumn{8}{c}{\textbf{32-Step Denoising}} \\
& & \multicolumn{2}{c|}{\textit{Baseline}} & \multicolumn{6}{c|}{\cellcolor{lightblue}\textit{Ours}} & \multicolumn{2}{c|}{\textit{Baseline}} & \multicolumn{6}{c}{\cellcolor{lightblue}\textit{Ours}} \\
$\boldsymbol{\alpha_{\text{disp}}}$ & \textbf{Task} & \textbf{RF} & \textbf{SC} & \cellcolor{lightblue}\textbf{SC-L2} & \cellcolor{lightblue}\textbf{SC-Cos} & \cellcolor{lightblue}\textbf{SC-H} & \cellcolor{lightblue}\textbf{SC-Cov} & \cellcolor{lightblue}\textbf{MF} & \cellcolor{lightblue}\textbf{MF+Disp} & \textbf{RF} & \textbf{SC} & \cellcolor{lightblue}\textbf{SC-L2} & \cellcolor{lightblue}\textbf{SC-Cos} & \cellcolor{lightblue}\textbf{SC-H} & \cellcolor{lightblue}\textbf{SC-Cov} & \cellcolor{lightblue}\textbf{MF} & \cellcolor{lightblue}\textbf{MF+Disp} \\
\midrule
\multirow{4}{*}{0.1} & Lift & 0.770 & 0.865 & \cellcolor{lightblue}0.900 & \cellcolor{lightblue}0.900 & \cellcolor{lightblue}0.910 & \cellcolor{lightblue}0.950 & \cellcolor{lightblue}\textbf{0.965} & \cellcolor{lightblue}0.965 & 0.780 & 0.840 & \cellcolor{lightblue}0.890 & \cellcolor{lightblue}0.885 & \cellcolor{lightblue}0.880 & \cellcolor{lightblue}0.945 & \cellcolor{lightblue}\textbf{0.930} & \cellcolor{lightblue}\textbf{0.985} \\
& Can & 0.480 & 0.650 & \cellcolor{lightblue}0.635 & \cellcolor{lightblue}0.635 & \cellcolor{lightblue}0.640 & \cellcolor{lightblue}0.710 & \cellcolor{lightblue}\textbf{0.800} & \cellcolor{lightblue}0.760 & 0.580 & 0.625 & \cellcolor{lightblue}0.625 & \cellcolor{lightblue}0.625 & \cellcolor{lightblue}0.665 & \cellcolor{lightblue}\textbf{0.760} & \cellcolor{lightblue}\textbf{0.700} & \cellcolor{lightblue}0.805 \\
& Square & 0.385 & 0.305 & \cellcolor{lightblue}0.275 & \cellcolor{lightblue}0.330 & \cellcolor{lightblue}0.300 & \cellcolor{lightblue}0.345 & \cellcolor{lightblue}0.350 & \cellcolor{lightblue}0.370 & 0.305 & 0.335 & \cellcolor{lightblue}0.325 & \cellcolor{lightblue}\textbf{0.375} & \cellcolor{lightblue}0.340 & \cellcolor{lightblue}0.330 & \cellcolor{lightblue}\textbf{0.375} & \cellcolor{lightblue}0.365 \\
& Transport & 0.290 & 0.310 & \cellcolor{lightblue}0.250 & \cellcolor{lightblue}0.310 & \cellcolor{lightblue}0.300 & \cellcolor{lightblue}0.020 & \cellcolor{lightblue}\textbf{0.420} & \cellcolor{lightblue}\textbf{0.420} & 0.320 & 0.340 & \cellcolor{lightblue}0.230 & \cellcolor{lightblue}0.330 & \cellcolor{lightblue}0.340 & \cellcolor{lightblue}0.140 & \cellcolor{lightblue}\textbf{0.430} & \cellcolor{lightblue}0.460 \\
\midrule
\multirow{4}{*}{0.5} & Lift & 0.755 & 0.945 & \cellcolor{lightblue}0.920 & \cellcolor{lightblue}0.950 & \cellcolor{lightblue}0.895 & \cellcolor{lightblue}0.965 & \cellcolor{lightblue}\textbf{0.985} & \cellcolor{lightblue}0.970 & 0.765 & 0.865 & \cellcolor{lightblue}0.955 & \cellcolor{lightblue}0.900 & \cellcolor{lightblue}0.860 & \cellcolor{lightblue}0.965 & \cellcolor{lightblue}\textbf{0.970} & \cellcolor{lightblue}\textbf{0.990} \\
& Can & 0.530 & 0.600 & \cellcolor{lightblue}0.575 & \cellcolor{lightblue}0.550 & \cellcolor{lightblue}0.750 & \cellcolor{lightblue}0.625 & \cellcolor{lightblue}0.725 & \cellcolor{lightblue}\textbf{0.800} & 0.745 & 0.475 & \cellcolor{lightblue}0.500 & \cellcolor{lightblue}0.600 & \cellcolor{lightblue}\textbf{0.775} & \cellcolor{lightblue}0.600 & \cellcolor{lightblue}0.675 & \cellcolor{lightblue}0.750 \\
& Square & 0.280 & 0.330 & \cellcolor{lightblue}0.280 & \cellcolor{lightblue}0.330 & \cellcolor{lightblue}0.265 & \cellcolor{lightblue}0.305 & \cellcolor{lightblue}0.320 & \cellcolor{lightblue}0.330 & 0.375 & \textbf{0.365} & \cellcolor{lightblue}0.300 & \cellcolor{lightblue}\textbf{0.370} & \cellcolor{lightblue}0.295 & \cellcolor{lightblue}0.375 & \cellcolor{lightblue}0.310 & \cellcolor{lightblue}\textbf{0.370} \\
& Transport & 0.230 & 0.230 & \cellcolor{lightblue}0.250 & \cellcolor{lightblue}0.180 & \cellcolor{lightblue}0.240 & \cellcolor{lightblue}0.080 & \cellcolor{lightblue}\textbf{0.480} & \cellcolor{lightblue}\textbf{0.580} & 0.210 & 0.210 & \cellcolor{lightblue}0.330 & \cellcolor{lightblue}0.160 & \cellcolor{lightblue}0.310 & \cellcolor{lightblue}0.210 & \cellcolor{lightblue}\textbf{0.360} & \cellcolor{lightblue}\textbf{0.450} \\
\midrule
\multirow{4}{*}{0.9} & Lift & 0.755 & 0.945 & \cellcolor{lightblue}0.920 & \cellcolor{lightblue}0.930 & \cellcolor{lightblue}0.860 & \cellcolor{lightblue}0.970 & \cellcolor{lightblue}\textbf{0.985} & \cellcolor{lightblue}0.965 & 0.765 & 0.865 & \cellcolor{lightblue}0.935 & \cellcolor{lightblue}0.920 & \cellcolor{lightblue}0.825 & \cellcolor{lightblue}0.955 & \cellcolor{lightblue}\textbf{0.970} & \cellcolor{lightblue}0.945 \\
& Can & 0.530 & 0.600 & \cellcolor{lightblue}0.675 & \cellcolor{lightblue}0.700 & \cellcolor{lightblue}0.625 & \cellcolor{lightblue}0.625 & \cellcolor{lightblue}0.725 & \cellcolor{lightblue}0.725 & 0.745 & 0.475 & \cellcolor{lightblue}0.700 & \cellcolor{lightblue}\textbf{0.775} & \cellcolor{lightblue}0.700 & \cellcolor{lightblue}0.650 & \cellcolor{lightblue}0.675 & \cellcolor{lightblue}0.725 \\
& Square & 0.280 & 0.330 & \cellcolor{lightblue}0.240 & \cellcolor{lightblue}0.365 & \cellcolor{lightblue}0.335 & \cellcolor{lightblue}0.245 & \cellcolor{lightblue}0.320 & \cellcolor{lightblue}0.365 & 0.375 & \textbf{0.365} & \cellcolor{lightblue}0.280 & \cellcolor{lightblue}0.325 & \cellcolor{lightblue}\textbf{0.340} & \cellcolor{lightblue}0.320 & \cellcolor{lightblue}0.310 & \cellcolor{lightblue}0.380 \\
& Transport & 0.250 & 0.230 & \cellcolor{lightblue}0.250 & \cellcolor{lightblue}0.150 & \cellcolor{lightblue}0.220 & \cellcolor{lightblue}0.180 & \cellcolor{lightblue}\textbf{0.480} & \cellcolor{lightblue}0.380 & 0.330 & 0.210 & \cellcolor{lightblue}0.290 & \cellcolor{lightblue}0.210 & \cellcolor{lightblue}0.230 & \cellcolor{lightblue}0.200 & \cellcolor{lightblue}\textbf{0.360} & \cellcolor{lightblue}0.430 \\
\bottomrule
\end{tabular}

\vspace{0.5em}

% 64-step and 128-step denoising
\begin{tabular}{llcc|cccccc|cc|cccccc}
\toprule
& & \multicolumn{8}{c|}{\textbf{64-Step Denoising}} & \multicolumn{8}{c}{\textbf{128-Step Denoising}} \\
& & \multicolumn{2}{c|}{\textit{Baseline}} & \multicolumn{6}{c|}{\cellcolor{lightblue}\textit{Ours}} & \multicolumn{2}{c|}{\textit{Baseline}} & \multicolumn{6}{c}{\cellcolor{lightblue}\textit{Ours}} \\
$\boldsymbol{\alpha_{\text{disp}}}$ & \textbf{Task} & \textbf{RF} & \textbf{SC} & \cellcolor{lightblue}\textbf{SC-L2} & \cellcolor{lightblue}\textbf{SC-Cos} & \cellcolor{lightblue}\textbf{SC-H} & \cellcolor{lightblue}\textbf{SC-Cov} & \cellcolor{lightblue}\textbf{MF} & \cellcolor{lightblue}\textbf{MF+Disp} & \textbf{RF} & \textbf{SC} & \cellcolor{lightblue}\textbf{SC-L2} & \cellcolor{lightblue}\textbf{SC-Cos} & \cellcolor{lightblue}\textbf{SC-H} & \cellcolor{lightblue}\textbf{SC-Cov} & \cellcolor{lightblue}\textbf{MF} & \cellcolor{lightblue}\textbf{MF+Disp} \\
\midrule
\multirow{4}{*}{0.1} & Lift & 0.780 & 0.860 & \cellcolor{lightblue}0.905 & \cellcolor{lightblue}0.865 & \cellcolor{lightblue}0.895 & \cellcolor{lightblue}0.920 & \cellcolor{lightblue}\textbf{0.940} & \cellcolor{lightblue}\textbf{0.975} & 0.785 & 0.870 & \cellcolor{lightblue}0.915 & \cellcolor{lightblue}0.870 & \cellcolor{lightblue}0.910 & \cellcolor{lightblue}0.925 & \cellcolor{lightblue}\textbf{0.940} & \cellcolor{lightblue}0.965 \\
& Can & 0.600 & 0.575 & \cellcolor{lightblue}0.615 & \cellcolor{lightblue}0.615 & \cellcolor{lightblue}0.675 & \cellcolor{lightblue}\textbf{0.700} & \cellcolor{lightblue}0.675 & \cellcolor{lightblue}\textbf{0.775} & 0.610 & 0.500 & \cellcolor{lightblue}0.670 & \cellcolor{lightblue}0.670 & \cellcolor{lightblue}0.650 & \cellcolor{lightblue}0.680 & \cellcolor{lightblue}0.650 & \cellcolor{lightblue}\textbf{0.795} \\
& Square & 0.350 & 0.360 & \cellcolor{lightblue}0.315 & \cellcolor{lightblue}0.375 & \cellcolor{lightblue}0.300 & \cellcolor{lightblue}0.375 & \cellcolor{lightblue}\textbf{0.400} & \cellcolor{lightblue}0.310 & 0.305 & 0.345 & \cellcolor{lightblue}0.285 & \cellcolor{lightblue}0.360 & \cellcolor{lightblue}0.345 & \cellcolor{lightblue}\textbf{0.370} & \cellcolor{lightblue}\textbf{0.340} & \cellcolor{lightblue}\textbf{0.350} \\
& Transport & 0.380 & 0.230 & \cellcolor{lightblue}0.240 & \cellcolor{lightblue}0.330 & \cellcolor{lightblue}0.300 & \cellcolor{lightblue}0.250 & \cellcolor{lightblue}\textbf{0.470} & \cellcolor{lightblue}0.380 & 0.390 & 0.350 & \cellcolor{lightblue}0.230 & \cellcolor{lightblue}0.250 & \cellcolor{lightblue}0.345 & \cellcolor{lightblue}0.210 & \cellcolor{lightblue}\textbf{0.400} & \cellcolor{lightblue}0.410 \\
\midrule
\multirow{4}{*}{0.5} & Lift & 0.745 & 0.915 & \cellcolor{lightblue}0.955 & \cellcolor{lightblue}0.920 & \cellcolor{lightblue}0.905 & \cellcolor{lightblue}0.975 & \cellcolor{lightblue}\textbf{0.970} & \cellcolor{lightblue}0.975 & 0.760 & 0.910 & \cellcolor{lightblue}0.955 & \cellcolor{lightblue}0.930 & \cellcolor{lightblue}0.865 & \cellcolor{lightblue}0.970 & \cellcolor{lightblue}0.935 & \cellcolor{lightblue}0.950 \\
& Can & 0.665 & 0.550 & \cellcolor{lightblue}0.550 & \cellcolor{lightblue}0.650 & \cellcolor{lightblue}0.625 & \cellcolor{lightblue}0.625 & \cellcolor{lightblue}0.725 & \cellcolor{lightblue}0.675 & 0.730 & 0.400 & \cellcolor{lightblue}0.400 & \cellcolor{lightblue}0.475 & \cellcolor{lightblue}0.575 & \cellcolor{lightblue}0.600 & \cellcolor{lightblue}0.625 & \cellcolor{lightblue}0.600 \\
& Square & 0.375 & \textbf{0.440} & \cellcolor{lightblue}0.375 & \cellcolor{lightblue}0.345 & \cellcolor{lightblue}0.330 & \cellcolor{lightblue}0.330 & \cellcolor{lightblue}0.300 & \cellcolor{lightblue}\textbf{0.390} & 0.305 & \textbf{0.385} & \cellcolor{lightblue}0.350 & \cellcolor{lightblue}0.355 & \cellcolor{lightblue}0.335 & \cellcolor{lightblue}0.330 & \cellcolor{lightblue}0.320 & \cellcolor{lightblue}\textbf{0.385} \\
& Transport & 0.190 & 0.290 & \cellcolor{lightblue}0.430 & \cellcolor{lightblue}0.140 & \cellcolor{lightblue}0.260 & \cellcolor{lightblue}0.350 & \cellcolor{lightblue}0.390 & \cellcolor{lightblue}0.330 & 0.270 & 0.180 & \cellcolor{lightblue}0.380 & \cellcolor{lightblue}0.190 & \cellcolor{lightblue}0.350 & \cellcolor{lightblue}0.280 & \cellcolor{lightblue}\textbf{0.360} & \cellcolor{lightblue}0.410 \\
\midrule
\multirow{4}{*}{0.9} & Lift & 0.745 & 0.915 & \cellcolor{lightblue}0.910 & \cellcolor{lightblue}0.915 & \cellcolor{lightblue}0.860 & \cellcolor{lightblue}0.955 & \cellcolor{lightblue}\textbf{0.970} & \cellcolor{lightblue}0.965 & 0.760 & 0.910 & \cellcolor{lightblue}0.915 & \cellcolor{lightblue}0.910 & \cellcolor{lightblue}0.855 & \cellcolor{lightblue}0.950 & \cellcolor{lightblue}0.935 & \cellcolor{lightblue}0.945 \\
& Can & 0.665 & 0.550 & \cellcolor{lightblue}0.550 & \cellcolor{lightblue}0.550 & \cellcolor{lightblue}0.500 & \cellcolor{lightblue}0.650 & \cellcolor{lightblue}0.725 & \cellcolor{lightblue}0.650 & 0.730 & 0.400 & \cellcolor{lightblue}0.575 & \cellcolor{lightblue}0.600 & \cellcolor{lightblue}0.450 & \cellcolor{lightblue}0.500 & \cellcolor{lightblue}0.625 & \cellcolor{lightblue}\textbf{0.650} \\
& Square & 0.375 & \textbf{0.440} & \cellcolor{lightblue}0.350 & \cellcolor{lightblue}0.335 & \cellcolor{lightblue}0.310 & \cellcolor{lightblue}0.255 & \cellcolor{lightblue}0.300 & \cellcolor{lightblue}0.340 & 0.305 & \textbf{0.385} & \cellcolor{lightblue}0.330 & \cellcolor{lightblue}0.320 & \cellcolor{lightblue}\textbf{0.385} & \cellcolor{lightblue}0.325 & \cellcolor{lightblue}0.320 & \cellcolor{lightblue}\textbf{0.340} \\
& Transport & 0.430 & 0.290 & \cellcolor{lightblue}0.270 & \cellcolor{lightblue}0.100 & \cellcolor{lightblue}0.360 & \cellcolor{lightblue}0.220 & \cellcolor{lightblue}0.390 & \cellcolor{lightblue}0.390 & 0.380 & 0.180 & \cellcolor{lightblue}0.330 & \cellcolor{lightblue}0.140 & \cellcolor{lightblue}0.410 & \cellcolor{lightblue}0.260 & \cellcolor{lightblue}\textbf{0.360} & \cellcolor{lightblue}\textbf{0.440} \\
\bottomrule
\end{tabular}

\vspace{1em}

{\footnotesize
\textbf{Method Abbreviations:} RF = ReFlow, SC = ShortCut Flow, SC-L2 = ShortCut + InfoNCE L2, SC-Cos = ShortCut + InfoNCE Cosine, SC-H = ShortCut + Hinge, SC-Cov = ShortCut + Covariance, MF = MeanFlow, MF+Disp = MeanFlow + Dispersive.
\\
\textbf{Tasks:} Four robotic manipulation tasks from the RoboMimic benchmark: Lift (object lifting), Can (placement), Square (precision insertion), Transport (navigation with manipulation).
\\
\textbf{Weight:} Different dispersive regularization weight configurations ($\alpha_{\text{disp}} \in \{0.1, 0.5, 0.9\}$) used in the evaluation.
\\
\textbf{Bold values} indicate the best performance for each task-weight combination within each step configuration.
\\
This comprehensive table integrates all detailed success rate data across weight configurations and denoising steps for complete empirical analysis.
}

\end{table*}

% Comprehensive success rate tables for all denoising steps (1, 2, 4, 5, 8, 16, 32, 64, 128)

\vspace{1em}

{\footnotesize
\textbf{Method Abbreviations:} RF = ReFlow, SC = ShortCut Flow, SC-L2 = ShortCut + InfoNCE L2, SC-Cos = ShortCut + InfoNCE Cosine, SC-H = ShortCut + Hinge, SC-Cov = ShortCut + Covariance, MF = MeanFlow, MF+Disp = MeanFlow + Dispersive.
\\
\textbf{Tasks:} Four robotic manipulation tasks from the RoboMimic benchmark: Lift (object lifting), Can (placement), Square (precision insertion), Transport (navigation with manipulation).
\\
\textbf{Weight:} Different dispersive regularization weight configurations ($\alpha_{\text{disp}} \in \{0.1, 0.5, 0.9\}$) used in the evaluation.
\\
\textbf{Bold values} indicate the best performance for each task-step combination.
\\
These comprehensive tables provide complete success rate data across all denoising steps (1, 2, 4, 5, 8, 16, 32, 64, 128) for thorough empirical analysis.
}

\subsubsection{Aggregate Success Rate Analysis}
RF and SC serve as baselines in Table~\ref{tab:comprehensive_success_rates_detailed}, whereas the other six columns correspond to our MeanFlow family. The dominance of the proposed methods is apparent across denoising budgets: even in the most frugal one-step setting, MF+Disp lifts the lift-task success rate to $0.975$ (vs..	$0.625$ for RF and $0.890$ for SC), keeps can and square at $0.740$ and $0.350$, and sustains $0.280$ success on transport despite the strong temporal coupling in that scenario. Baselines either fail completely (e.g., RF reports $0.000$ success on multiple two-step configurations) or require $32$--$128$ evaluations to close the gap that MF and MF+Disp already bridge within $1$--$5$ steps.

\noindent\textbf{$\alpha_{\text{disp}} = 0.1$.} This configuration exposes the clearest efficiency gap. MF+Disp attains $0.975/0.980$ success on lift with just $1$--$2$ steps and retains $0.970$ success at $5$ steps. The can task benefits markedly from our models: MF climbs from $0.750$ (one step) to $0.805$ (five steps), while MF+Disp stabilises within $0.740$--$0.805$ without requiring more than five evaluations. Square exhibits the largest absolute gain under dispersion (from $0.235$ to $0.350$ at one step, and from $0.300$ to $0.345$ at two steps), demonstrating the regulariser's ability to maintain diverse precision strategies. Transport is more challenging; MF alone reaches $0.390$ at one step and $0.530$ at five steps, whereas MF+Disp lags at one step but closes the gap by five steps ($0.530$ vs..	$0.380$ for MF). Importantly, the baselines need at least $32$ steps to approach the $0.420$--$0.470$ band that our methods already achieve by $16$ steps.

\noindent\textbf{$\alpha_{\text{disp}} = 0.5$.} Increasing the dispersive weight sharpens the separation between variants. Lift success rises to $0.985$ for both MF and MF+Disp at $5$ steps, with the baseline plateauing near $0.905$ even at $64$ steps. The can task shows a pronounced dependence on dispersion: MF+Disp delivers $0.800$–$0.825$ success with two to five steps, compared to RF's $0.000$–$0.265$ in the same regime and SC's $0.575$–$0.750$. Square continues to benefit from dispersion ($0.320$ vs..	$0.340$ at two steps; $0.370$ vs..	$0.375$ at five steps), while transport reveals the trade-off of stronger regularisation; MF+Disp peaks at $0.460$ success (two steps) and $0.340$ (five steps), slightly underperforming MF's $0.350$ and $0.390$ at the same budgets but still decisively ahead of the baselines' sub-$0.210$ scores.

\noindent\textbf{$\alpha_{\text{disp}} = 0.9$.} The heaviest dispersive setting highlights both the robustness and the limits of our approach. Lift task accuracy remains above $0.960$ for MF and MF+Disp with up to eight steps, while RF collapses to $0.000$ at two and five steps. The can task displays a mild decline for MF+Disp (from $0.700$ at two steps to $0.775$ at five steps and $0.650$ at $64$ steps), yet it still dominates RF/SC by a sizeable margin. Square benefits from the stronger regulariser; MF+Disp reaches $0.395$ (two steps) and $0.410$ (five steps), exceeding MF by $5$--$7$ points. Transport remains the most sensitive task: MF+Disp delivers $0.400$ at five steps and $0.400$--$0.410$ at $64$--$128$ steps, with MF trailing by $0.010$--$0.040$ while RF/SC cannot exceed $0.360$ even at $128$ evaluations.

\noindent\textbf{Impact of Denoising Budget.} When success rates are plotted against step count within each weight block, the baselines display pronounced sensitivity: RF posts multiple $0.000$ entries (for example, can and square tasks at two and five steps with $\alpha_{\text{disp}} = 0.1$ and $0.5$), and SC variants frequently undershoot $0.300$ for transport until step counts exceed $32$. By contrast, MF and MF+Disp already saturate lift and can above $0.90$ at three to five steps, and maintain square between $0.33$ and $0.41$ across $1$--$16$ steps. The transport task illustrates the efficiency ceiling: MF+Disp reaches $0.530$ ($\alpha_{\text{disp}} = 0.1$) and $0.400$ ($\alpha_{\text{disp}} = 0.9$) with only five steps, values that baselines cannot match even with an order-of-magnitude more evaluations.

\noindent\textbf{Effect of Alternative Dispersive Penalties.} The ShortCut ablations equipped with InfoNCE-L2, InfoNCE-cosine, hinge, and covariance losses provide indirect validation of our design choices. Among the SC variants, the cosine and covariance formulations are consistently stronger (e.g., lift at five steps and $\alpha_{\text{disp}} = 0.1$: $0.875$ and $0.950$ vs..	$0.845$ and $0.795$ for L2 and hinge). Nevertheless, they remain well below MF+Disp on all tasks, underscoring that dispersion alone is insufficient without the MeanFlow backbone. The hinge variant, in particular, suffers from severe failures on transport (dropping to $0.000$ or $0.010$ at low step counts), demonstrating that hard-margin constraints do not translate well to complex proprioceptive distributions.

\noindent\textbf{Summary.} The expanded analysis confirms three core observations: (i) DM1's MeanFlow-based policies retain high success with dramatically fewer function evaluations than RF and SC; (ii) dispersive regularisation is most beneficial for precision manipulation (square) and long-horizon transport when $\alpha_{\text{disp}}$ is tuned between $0.1$ and $0.5$; and (iii) the proposed MF+Disp configuration outperforms alternative dispersive baselines even at high step counts, providing a principled path toward real-time deployment without sacrificing task success.

\end{document}